\pdfoutput=1

\documentclass[acmsmall,screen]{acmart}

\usepackage{amsthm}
\usepackage{tcolorbox}
\usepackage{tikz} 
\usetikzlibrary{shapes.geometric}

\usepackage{booktabs}
\usepackage{graphicx}
\newtcolorbox{reflectionbox}{
  colback=blue!5!white,
  colframe=blue!50!black,
  fonttitle=\bfseries,
  sharp corners,
  boxrule=0.5pt,
  left=6pt,
  right=6pt,
  top=6pt,
  bottom=6pt
}

\usepackage{array}
\newcolumntype{Y}{>{\raggedright\arraybackslash}X}

\usepackage{tabularx}

\newtcolorbox{rqbox}[1]{title={#1}, fonttitle=\bfseries}
\newtcolorbox{evidencebox}[1]{title={#1}, fonttitle=\bfseries}
\newtcolorbox{takeawaybox}[1]{title={#1}, fonttitle=\bfseries}

 \usepackage{mdframed}
\newmdenv[
  skipabove=8pt, skipbelow=8pt,
  linewidth=0.8pt, roundcorner=3pt,
  innerleftmargin=8pt, innerrightmargin=8pt,
  innertopmargin=6pt, innerbottommargin=6pt
]{keybox}

\newcounter{principle}
\renewcommand{\theprinciple}{\arabic{principle}} 

\newenvironment{principle}{%
  \refstepcounter{principle}%
  \par\addvspace{0.6\baselineskip}%
  \noindent\itshape\textbf{Principle \theprinciple. }\ignorespaces%
}{%
  \par\addvspace{0.6\baselineskip}%
}

\AtBeginDocument{%
  }

\begin{document}

\title{
    The ARC of Progress towards AGI: A Living Survey of Abstraction and Reasoning
}

\author{Sahar Vahdati}
\email{Sahar.Vahdati@tib.eu}
\orcid{0000-0002-7171-169X}
\affiliation{
  \institution{TIB - Leibniz Information Centre for Science and Technology}
  \city{Hannover}
  \country{Germany}
}
\affiliation{
  \institution{L3S Research Center, Leibniz University of Hannover}
  \city{Hannover}
  \country{Germany}
}

\author{Andrei Aioanei}
\email{Andrei.Aioanei@tib.eu}
\orcid{0000-0001-5547-9969}
\affiliation{
  \institution{TIB - Leibniz Information Centre for Science and Technology}
  \city{Hannover}
  \country{Germany}
}

\author{Haridhra Suresh}
\email{Haridhra.Suresh@tib.eu}
\orcid{0009-0005-2078-2956}
\affiliation{
  \institution{TIB - Leibniz Information Centre for Science and Technology}
  \city{Hannover}
  \country{Germany}
}

\author{Jens Lehmann}
\email{jlehmnn@amazon.de}
\orcid{0000-0001-9108-4278}
\affiliation{
  \institution{Dresden University of Technology, Amazon (work done outside of Amazon)}
  \city{Dresden}
  \country{Germany}
}

\renewcommand{\shortauthors}{Vahdati et al.}

\begin{abstract}
    The Abstraction and Reasoning Corpus (ARC-AGI) has become a key
    benchmark for fluid intelligence in AI. This survey presents the first
    cross-generation analysis of 82 approaches across three benchmark
    versions and the ARC Prize 2024-2025 competitions.
    Our central finding is that performance degradation across versions is
    consistent across all paradigms: program synthesis, neuro-symbolic, and
    neural approaches all exhibit 2-3$\times$ drops from ARC-AGI-1 to
    ARC-AGI-2, indicating fundamental limitations in compositional
    generalization. While systems now reach 93.0\% on ARC-AGI-1
    (Opus~4.6), performance falls to 68.8\% on ARC-AGI-2 and 13\% on
    ARC-AGI-3, as humans maintain near-perfect accuracy across all versions.
    Cost fell 390$\times$ in one year (o3's \$4,500/task to GPT-5.2's
    \$12/task), although this largely reflects reduced test-time parallelism.
    Trillion-scale models vary widely in score and cost, while
    Kaggle-constrained entries (660M-8B) achieve competitive results,
    aligning with Chollet's thesis that intelligence is skill-acquisition
    efficiency~\cite{chollet_measure_2019}.
    Test-time adaptation and refinement loops emerge as critical success
    factors, while compositional reasoning and interactive learning remain
    unsolved. ARC Prize 2025 winners needed hundreds of thousands of
    synthetic examples to reach 24\% on ARC-AGI-2, confirming that
    reasoning remains knowledge-bound.
    This first release of the \textbf{ARC-AGI Living Survey} captures the
    field as of February 2026, with updates at
    \url{https://nimi-ai.com/arc-survey/}.
\end{abstract}



\maketitle

%

\section{Introduction}\label{sec:introduction}

The pursuit of Artificial General Intelligence (AGI) requires benchmarks
that can distinguish genuine reasoning from sophisticated pattern matching.
The Abstraction and Reasoning Corpus (ARC-AGI)~\cite{chollet_measure_2019}
stands out as such a benchmark, highlighting important limitations about the current
state of AI.
While humans solve these tasks with near-perfect accuracy,
AI systems show dramatic performance variation. 
Since its 2019 introduction,
ARC-AGI has evolved through three progressively more challenging versions.
The tasks are few-shot grid-transformation problems. Given a number of input-output example grids (typically 3-5), a solver (system or human) must infer the rule and produce the exact output grid for a held-out test input.
To reduce benchmark-specific tuning and contamination, evaluations use a public development/feedback set and a disjoint private (hidden) set for official leaderboard and competition scoring.
On ARC-AGI-1, frontier models now exceed 96\% accuracy (Gemini~3 Deep Think
at \$7.17/task), with Opus~4.6 at 93.0\% (\$1.88/task) and GPT-5.2~Pro at
90.5\% (\$11.64/task)~\cite{arcprize2025_leaderboard}, the latter representing
a 390$\times$ efficiency improvement in one year from o3's \$4,500/task.
However, performance drops on ARC-AGI-2.
Gemini~3 Deep Think leads at 84.6\%
(\$13.62/task), while under Kaggle resource constraints the best system (NVARC)
achieved only 24\%~\cite{arcprize2025_results}. 
On ARC-AGI-3, performance drops further to 13\%~\cite{arcagi3_learning}.
While ARC Prize designed the benchmark to be fully solvable by humans,
evaluation studies show that each ARC-AGI-2 task is solved by roughly 75\%
of individual participants~\cite{chollet_arc-agi-2_2025}, reflecting
task difficulty variation rather than fundamental human limitations.
NYU published a study assessing Mechanical Turk workers, which showed
an average worker solves 77\% of the ARC-AGI-1 public evaluation tasks
\cite{legris_h-arc_2024, arcprize2025_leaderboard}.

The transformation from near-zero to near-human performance on ARC-AGI-1 came
from a paradigm shift away from massive pre-training and toward test-time compute.
Some systems do this via program synthesis, while others primarily increase
inference-time reasoning effort or sampling. 
This breakthrough aligns with
Chollet's hypothesis that achieving AGI requires not larger models but different
architectures capable of efficient skill acquisition and compositional generalization
~\cite{chollet_arc_2025, akyurek_surprising_2025}.
A central theme emerging from the ARC Prize 2025 is that \emph{refinement is
intelligence}, systems explore candidate solutions, verify results through
feedback signals, and iterate through refinement loops until
convergence~\cite{arcprize2025_results}. 
The paper award winners exemplified
this: the Tiny Recursive Model (TRM) by Jolicoeur-Martineau achieved 45\% on
ARC-AGI-1 with only 7M parameters through recursive latent
refinement~\cite{jolicoeur2025_trm}, while CompressARC by Liao and
Gu reached 20--34\% with merely 76K parameters using MDL-based compression~\cite{liao_arc-agi_2025}. 
Both approaches demonstrate that test-time training on individual puzzles, rather than massive pretraining, enables efficient reasoning on novel tasks.
We base the following sections on the question of:

\begin{reflectionbox}
\small
\setlength{\parskip}{0pt}
\setlength{\baselineskip}{0.95\baselineskip}
Can computational processes, however sophisticated, capture the fluid intelligence
that allows humans to rapidly adapt to novel domains with minimal examples
and resources?
\end{reflectionbox}

%


\subsection{The Intelligence Measurement Problem}\label{subsec:the-measurement-problem}

Current AI evaluations predominantly measure \textit{crystallized intelligence},
the ability to apply memorized skills to familiar problems. Chollet argued that
benchmarks prevalent in 2019, such as those testing domain knowledge and pattern
recognition, could be solved through sufficient scale without genuine
understanding~\cite{chollet_measure_2019}. This observation extends to recent
benchmarks: breakthroughs on MMLU, HLE, and mathematical olympiads have been
achieved through scaling rather than novel reasoning mechanisms~\cite{phan2025humanitysexam}.
Models can achieve PhD-level performance in specialized domains while failing
at children's puzzles, revealing that reasoning becomes entangled with
domain-specific knowledge rather than emerging as a transferable capability~\cite{berman2025-substack}.

ARC-AGI addresses this by measuring \textit{fluid intelligence}, the ability
to solve novel problems using minimal prior knowledge and few examples. Each
ARC task presents 3-5 input-output grid demonstrations and requires predicting
outputs for new inputs. Crucially, tasks are designed to be solvable by humans
without specialized knowledge, using only \textit{core knowledge priors}:
foundational cognitive abilities identified in developmental psychology, including
object permanence, goal-directedness, basic geometry, and numerosity~\cite{
chollet_measure_2019, chollet_arc_2025}.

Drawing on human cognitive tests like Raven's Progressive Matrices~\cite{raven_progressive_1938} and Elizabeth Spelke's core knowledge framework~\cite{spelke_core_2007}, Chollet's 2019 paper ``On the Measure of Intelligence'' argued that intelligence should be defined by a system's efficiency at acquiring skills in unfamiliar scenarios. He formalized this as the speed of learning new tasks given limited experience and innate priors~\cite{chollet_measure_2019}.
Recent computational analyses of such matrices~\cite{Yang2023ComputationalMO} further validate their utility for probing abstract reasoning.

\textbf{Evolution Across Three Generations.}
The ARC benchmark has evolved through three distinct versions, each addressing
limitations discovered in its predecessor while maintaining the core focus on
fluid intelligence.

\begin{figure}[t!]
    \centering
    \includegraphics[width=\textwidth, trim=0.5cm 23.2cm 0.5cm 0.5cm, clip]{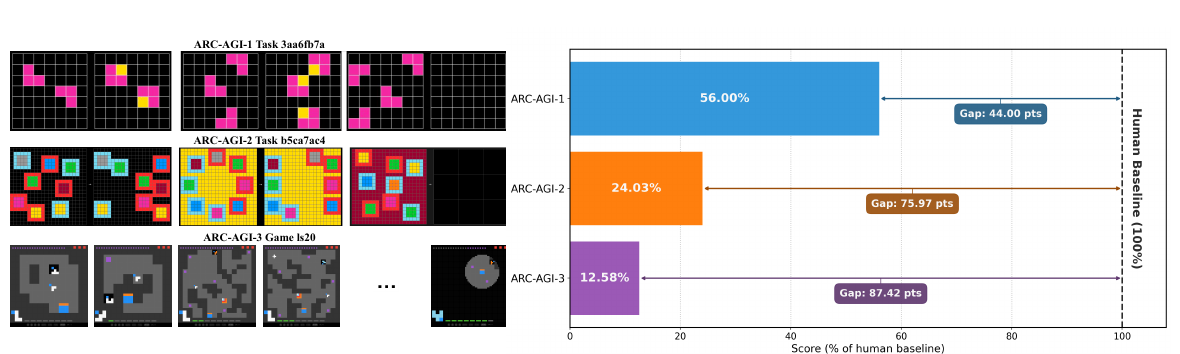}
    \caption{
        \textbf{Example task from each ARC-AGI version and the performance cliff across them.}
        On the left side, we show one example from each ARC-AGI version, illustrating
        the increased complexity and requirements. On the right side, horizontal stacked
        bars show best AI performance (darker color) and the gap to human baseline
        (lighter color) for each benchmark version.
    }
    \label{fig:performance-cliff}
\end{figure}

\noindent\textbf{ARC-AGI-1 (November 2019):} The original corpus of 1{,}000
program-induction tasks (400 training, 400 evaluation, 200 test) established
the benchmark~\cite{chollet_measure_2019}. Each task requires inferring a
transformation rule from around three input--output demonstrations. While
initially out of reach for AI systems, 2024 methods showed that, given massive
test-time compute, the benchmark can be solved to a substantial degree.

\noindent\textbf{ARC-AGI-2 (March 2025):} Announced in 2022 as part of the
benchmark's planned evolution, ARC-AGI-2 preserves the ARC format but sharply
increases difficulty via deeper multi-step compositionality, richer symbolic
interpretation, context-dependent rule application, and explicit resistance to
brute-force search~\cite{chollet_arc-agi-2_2025}. Frontier models now reach
84.6\% on the public leaderboard (Gemini~3 Deep Think at \$13.62/task), while
under Kaggle resource constraints the best system (NVARC) achieved 24\% in the
private competition~\cite{arcprize2025_results, arcprize2025_leaderboard}.
Expert human panels achieve full completion.

\noindent\textbf{ARC-AGI-3 Preview (July 2025):} ARC-AGI-3 reimagines evaluation
as interactive games in which agents must discover goals and mechanics
through exploration~\cite{arcagi3_learning}. The preview released six games;
the best AI system (StochasticGoose) achieved only 12.58\% action efficiency,
while over 1,200 human players completed more than 3,900 games, most successfully.
The full benchmark (1,000+ levels across 150+ environments) launches in March
2026. This gap highlights a shift from ``intelligence as pattern matching'' on
static datasets to ``intelligence as adaptive behavior'' in open-ended
environments~\cite{Ying2025-WorldModels}.

\subsection{Two Axes of Fragility: Performance Cliffs and Computational Efficiency}
\label{subsec:fundamental-limits}

A striking pattern emerges across ARC-AGI versions.
Each breakthrough on one
version has substantially lower performance on a higher version, see the right side of Figure~\ref{fig:performance-cliff}.
This performance cliff is particularly significant because it occurs \textit{consistently
across all paradigms}. Program synthesis approaches, neuro-symbolic systems, and
pure neural methods all exhibit 2.5-3$\times$ performance degradation from one version
to the next. This cross-paradigm consistency indicates a shared fundamental
limitation in compositional reasoning capabilities rather than a weakness of
any particular architectural choice.

The cliff stands in contrast to other AI benchmarks where scaling compute and
model size reliably improves performance until saturation. While larger models
do score higher within each ARC-AGI version, we cannot determine whether this
reflects genuine reasoning improvements or data contamination and benchmark-specific
optimization, precisely the confound ARC-AGI was designed to expose. The critical
diagnostic is cross-version transfer: frontier models narrow the
ARC-AGI-1 to ARC-AGI-2 gap only through orders-of-magnitude cost
increases, while under Kaggle resource constraints the best systems
drop to 16--24\% on ARC-AGI-2, suggesting that current improvements
reflect computational investment rather than compositional generalization.

\begin{reflectionbox}
   When frontier models maintain 85\% across ARC-AGI versions but only by
   spending \$7--14 per task, while resource-constrained systems drop to
   16--24\% on ARC-AGI-2, is the field closing the reasoning gap or merely
   masking it with scale?
\end{reflectionbox}

\noindent Section~\ref{sec:empirical-analysis} analyzes this question in detail,
examining which architectural components enable partial success and which
fundamental capabilities remain absent.

\noindent\textbf{The Efficiency Challenge.} These compositional limitations are mirrored
along a second axis: current systems incur extreme computational costs when they do
perform well. The benchmark explicitly treats computational cost as part of the
capability signal, recognizing that human intelligence solves these tasks with
minimal resources~\cite{chollet_arc_2025}. Cost per task spans five orders of
magnitude across systems, and even the most efficient frontier models remain
20--40$\times$ more expensive than human cognitive effort
(Section~\ref{subsec:cost-performance-frontiers}). Meanwhile, parameter-efficient
approaches such as TRM (7M parameters) and CompressARC (76K parameters) demonstrate
that competitive performance need not require massive scale~\cite{jolicoeur2025_trm,
liao_arc-agi_2025}.

\begin{reflectionbox}
 Does achieving high performance through massive test-time search constitute genuine
intelligence, or merely brute-force approximation of intelligent behavior?
\end{reflectionbox}

\noindent We examine this trade-off in detail, revealing that algorithmic innovation, particularly
refinement loops and test-time adaptation, provides far greater efficiency gains than
scaling compute alone.

\subsection{Structure}
\label{subsec:contributions}
This survey provides a comprehensive review of the ARC-AGI benchmark as of December 2025,
drawing from 80 papers collected through systematic search, of which 65 include rigorous
evaluations on standardized test sets (see Supplementary Material, Section~2, for
selection criteria). We examine the benchmark's design principles, task taxonomy, and
three-generation evolution (Section~\ref{sec:arc-agi-benchmark}).
Comparative analysis
identifies success factors, failure modes, and the performance cliff across benchmark
versions (Section~\ref{sec:comparative-analysis}), while empirical analysis traces
performance evolution, cost-efficiency trade-offs, and evaluation
rigor (Section~\ref{sec:empirical-analysis}). The discussion synthesizes findings through
three lenses: intelligence measurement, the compression paradox, and remaining challenges
on the path to AGI (Section~\ref{sec:discussion-and-implications}). We introduce the
\textbf{ARC-AGI Living Survey\footnote{\url{https://nimi-ai.com/survey/}}}, a continuously
updated repository that tracks the evolving landscape of ARC-AGI approaches, results,
and evaluation practices, preserving prior analyses as versioned snapshots while
integrating new developments.
This paper constitutes its first release.

\section{The ARC-AGI Benchmark}\label{sec:arc-agi-benchmark}

This section examines the ARC-AGI benchmark's design principles, task structure,
and evolution across three versions. Understanding these technical details is essential
for analyzing why current AI systems struggle and what approaches show promise.

\subsection{Design Principles}\label{subsec:design-principles}

The Abstraction and Reasoning Corpus fundamentally differs from conventional AI benchmarks.
Rather than testing accumulated knowledge or pattern matching on large datasets, ARC-AGI
measures \emph{fluid intelligence}, the capacity to efficiently acquire new skills and
solve novel problems with minimal prior knowledge~\cite{chollet_measure_2019}. The benchmark
grounds evaluation in a minimal set of innate cognitive abilities identified in developmental
psychology (objectness, agentness, numerosity, and basic geometry) while avoiding
domain-specific knowledge from mathematics, language, or culture~\cite{spelke_core_2007}.
This design creates a ``culture-fair'' test where human and machine intelligence can be
compared on equal footing.

Each ARC task presents 3-5 input-output grid pairs demonstrating a transformation rule,
which must be inferred and applied to novel test cases. Grids are small (typically under
30$\times$30 cells) with limited color palettes (10 colors) to minimize perceptual difficulty
while maximizing reasoning demands. The evaluation allows two attempts per task, reflecting
human problem-solving where initial hypotheses are often refined. Performance measures the
percentage of tasks solved with exact match; partial credit is not awarded, as in many
real-world scenarios approximately correct is insufficient~\cite{chollet_measure_2019}.
Crucially, the benchmark emphasizes \emph{developer-aware generalization}: test tasks
intentionally differ from any training data, preventing solutions based on memorization
or statistical pattern matching~\cite{chollet_measure_2019}. Additionally, computational
efficiency is treated as part of the capability signal; systems achieving high accuracy
only through massive computational expense are not considered to demonstrate human-like
intelligence~\cite{chollet_arc_2025}.

\subsection{Task Structure and Cognitive Taxonomy}\label{subsec:task-taxonomy}

\begin{figure}[!t]
    \centering
    \includegraphics[width=0.6\textwidth]{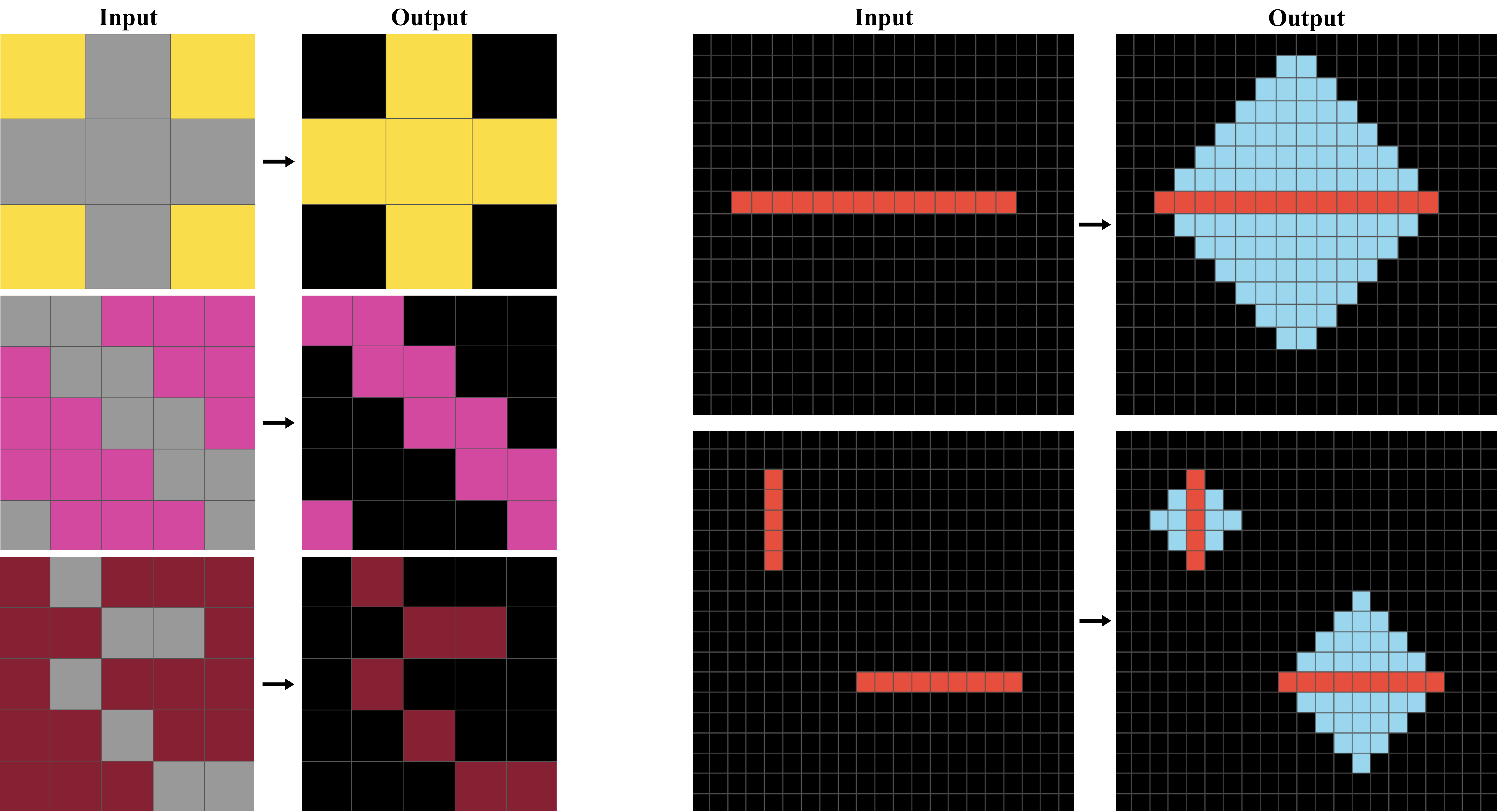}
    \caption{
        Representative ARC-AGI tasks illustrating two core reasoning categories.
        \textbf{Left:} Object-centric reasoning (Task f76d97a5, 3 training examples).
        The transformation extracts the colored checkerboard pattern from a gray
        background, requiring the system to identify ``object'' versus ``background''
        without explicit segmentation cues.
        \textbf{Right:} Geometric transformation (Task c97c0139, 2 training examples).
        Red line segments define reflection axes around which cyan diamond shapes
        must be generated symmetrically. Both tasks require inferring abstract rules
        from minimal demonstrations and generalizing to novel configurations.
    }
    \label{fig:arc-task-examples}
\end{figure}

ARC-AGI-1 and ARC-AGI-2 tasks (static, grid-based formats) probe six fundamental categories of abstract reasoning:
\textbf{object-centric reasoning} (identifying coherent objects, tracking properties, and applying transformations that respect boundaries---a challenge when objects are implicit or multiple segmentations are plausible),
\textbf{geometric transformations} (rotation, reflection, scaling, translation, and symmetry),
\textbf{relational and spatial reasoning} (containment, adjacency, alignment, and relative positioning),
\textbf{numerical reasoning} (counting, comparison, and using numbers to parameterize transformations),
and \textbf{pattern completion} (detecting repeating structure and extrapolating to extend or complete it).
Finally, \textbf{compositional reasoning}, the most challenging category, demands combining
multiple reasoning steps or applying several rules in sequence, testing the ability to
flexibly combine learned primitives and maintain intermediate representations. ARC-AGI-3
shifts to interactive environments requiring additional capabilities beyond these
categories, as discussed in Section~\ref{subsec:benchmark-evolution}.
Figure~\ref{fig:arc-task-examples} illustrates representative examples from these categories.

\subsection{Evolution Across Three Generations}\label{subsec:benchmark-evolution}

The ARC-AGI benchmark has evolved through three versions, each addressing limitations
revealed by progress on its predecessor while maintaining the core focus on fluid
intelligence. Table~\ref{tab:arc-versions-comparison} summarizes the key differences.

\begin{table}[!t]
    \caption{Comparison of ARC-AGI benchmark versions showing progressive increases in
    difficulty and the dramatic widening of the human-AI performance gap.}
    \label{tab:arc-versions-comparison}
    \centering
    \footnotesize
    \begin{tabular*}{\linewidth}{@{\extracolsep{\fill}}llll@{}}
        \toprule
        \textbf{Feature} & \textbf{ARC-AGI-1} & \textbf{ARC-AGI-2} & \textbf{ARC-AGI-3} \\
        \midrule
        Release date         & Nov 2019                & Mar 2025                 & Jul 2025 \\
        Format               & Static                  & Static                   & Interactive \\
        Number of tasks      & 1,000                   & 600                      & Variable \\
        Grid size            & $\leq$30$\times$30      & $\leq$30$\times$30       & 64$\times$64 \\
        Color palette        & 10 colors               & 10 colors                & 16 colors \\
        Best AI performance  & 90.5\%$^\dagger$ / 55.5\%$^\ddagger$
                            & 54.2\%$^\dagger$ / 24.03\%$^\ddagger$
                            & 12.58\% \\
        Human performance    & 100\%                   & 100\%                    & 100\% \\
        Performance gap      & 9.5\%                   & 46\%--76\%               & 87.42\% \\
        \bottomrule
    \end{tabular*}

    \vspace{2pt}
    \begin{tabular*}{\linewidth}{@{}l@{}}
        \footnotesize $^\dagger$Public leaderboard (GPT-5.2 Pro)\\
        \footnotesize $^\ddagger$Competition winner under resource constraints
        (ARC-AGI-1: MindsAI 55.5\%, 2024; ARC-AGI-2: NVARC 24.03\%, 2025)
    \end{tabular*}
\end{table}

\subsubsection{ARC-AGI-1: Establishing the Baseline}

The original corpus, released in 2019, comprises 1,000 unique tasks divided into 400
training, 400 evaluation (publicly available), and 200 private test tasks. Each task
contains 3--5 input-output demonstration pairs and typically one test input (occasionally two).
Tasks exhibit clear difficulty progression: simple uniform transformations (solved by
>90\% of systems), object-centric operations with geometric transformations (40-60\%
success rate), and multi-step reasoning requiring abstraction (<10\% success rate).

For five years, ARC-AGI-1 successfully resisted most AI approaches, with performance
remaining below 20\% until 2024. The ARC Prize 2024 competition, offering \$600,000
for surpassing 85\% accuracy, catalyzed rapid progress with over 1,430 teams competing.
The top open-source system (ARChitects) achieved 53.5\% through test-time training,
demonstrating that architectural innovations could break the long-standing performance
barrier~\cite{arcagi3_learning}. However, analysis of successful systems revealed
exploitable characteristics: limited compositional depth (most tasks require 1-2
reasoning steps), statistical regularities in transformation types, and vulnerability
to brute-force search given sufficient test-time compute. These findings motivated
ARC-AGI-2's design~\cite{chollet_arc-agi-2_2025}.

\subsubsection{ARC-AGI-2: Targeting Compositional Complexity}

Released in March 2025 following OpenAI o3's breakthrough on ARC-AGI-1, this version
maintains the same input-output format while significantly increasing difficulty through
four targeted enhancements~\cite{chollet_arc-agi-2_2025, arcprize2024_o3}. First, tasks require deeper
compositional reasoning through multi-step transformations where the state after step N
depends on step N-1, making it virtually impossible to predict outcomes without executing
sequential operations. 
Many tasks require symbolic, context-dependent pattern interpretation and rule application that adds control-flow complexity.
Careful construction expands search spaces and introduces subtle distinctions between correct and plausible-but-wrong solutions, resisting
brute-force enumeration~\cite{chollet_arc-agi-2_2025}.

The ARC Prize 2025 competition revealed the challenge's severity: NVARC achieved
24.03\% (1st place), The ARChitects 16.53\% (2nd), and MindsAI 12.64\%
(3rd), with 1,455 teams submitting 15,154 entries and 90 papers (up from 47
in 2024)~\cite{arcprize2025_results}. Commercial systems fared better: Opus 4.5
reached 37.6\% at \$2.20/task, while Poetiq's Gemini 3 Pro refinement approach
achieved 54\% at \$30/task through iterative program
transformation~\cite{arcprize2025_leaderboard}. Human evaluation studies with over 400 test-takers
found that while expert panels solve 100\% of tasks, average individual test-takers
score 60\%, with each task solved by approximately 75\% of those who attempted
it~\cite{chollet_arc-agi-2_2025}.
This 2.5--3$\times$ performance degradation across all AI
paradigms indicates
fundamental limitations in compositional generalization rather than paradigm-specific
weaknesses (Section~\ref{sec:empirical-analysis}).

A central theme from the 2025 competition was that \emph{refinement is intelligence}:
top-performing systems employ iterative refinement loops that explore candidate solutions,
verify results through feedback signals, and repeat until convergence~\cite{arcprize2025_results}.
The paper award winners exemplified parameter-efficient approaches: the Tiny Recursive
Model (TRM) by Jolicoeur-Martineau achieved 45\% on ARC-AGI-1 with only 7M parameters
through recursive latent refinement~\cite{jolicoeur2025_trm}, while CompressARC
by Liao and Gu reached 20--34\% with merely 76K parameters using MDL-based
compression~\cite{liao_arc-agi_2025}. 
Both demonstrate that test-time training on individual puzzles, where all task-specific learning happens at inference time, can outperform pretrained models.

\subsubsection{ARC-AGI-3: The Interactive Paradigm Shift}

The ARC-AGI-3 preview (July 2025) represents a fundamental reconceptualization of
intelligence evaluation~\cite{arcagi3_learning, Ying2025-WorldModels}. Rather than
presenting static input-output pairs, it challenges systems with interactive
mini-games where rules and goals must be discovered through autonomous exploration.
This shift addresses a critical limitation of static benchmarks: they test whether
systems can \emph{recognize} patterns but not whether they can \emph{discover} them
through active learning.

Each game provides a 64$\times$64 grid environment with a 16-color palette. Games
comprise sequential levels that enable transfer learning, and provide no
instructions: agents must explore to discover mechanics, infer goals from sparse
feedback, and plan multi-step action sequences. Scoring measures \emph{action
efficiency}, comparing how many actions an agent requires relative to human
baselines~\cite{arcagi3_learning}. This format tests exploration, memory,
hypothesis-driven experimentation, and meta-learning across games, capabilities
that align closely with how humans approach novel problems.

The preview released six games (three public, three private holdout). The top AI
system (StochasticGoose, CNN-based reinforcement learning) achieved 12.58\% action
efficiency, while over 1,200 human players completed more than 3,900 games, most
successfully~\cite{arcagi3_learning}. This gap, 8$\times$ larger than on ARC-AGI-1,
reveals that even systems performing well on static reasoning lack the autonomous
learning capabilities central to human intelligence. The full benchmark (1,000+
levels across 150+ environments) launches in March 2026.



\begin{rqbox}{Implications for Intelligence Measurement.}
    The benchmark's three-generation evolution highlights three core principles.
    \vspace{-0.2cm}
    \noindent
    \begin{principle}
        The compositional cliff persists but is scale-dependent.
    \end{principle}
    \vspace{-0.2cm}
    \noindent Frontier models are narrowing the ARC-AGI-1 to ARC-AGI-2 gap:
    Gemini~3 Deep Think drops from 96.0\% to 84.6\% ($-$12\%) and
    Opus~4.6 from 93.0\% to 68.8\% ($-$26\%). Yet under Kaggle resource
    constraints, the cliff remains severe: NVARC achieves 24\% and the
    ARChitects 16.5\% on ARC-AGI-2~\cite{arcprize2025_results}. The
    30--60 percentage point gap between unconstrained and constrained
    regimes suggests that current progress reflects computational
    investment as much as architectural insight.

    \begin{principle}
        Accuracy is converging; efficiency and transparency are not.
    \end{principle}
    \vspace{-0.2cm}
    \noindent Public leaderboard systems approach human baselines on
    ARC-AGI-2, but these are proprietary models whose training corpora,
    synthetic data pipelines, scale, and potential benchmark exposure
    remain opaque~\cite{chollet_arc_2025}. 
    ARC leaderboard gains may mostly come from massive synthetic-data coverage, not true reasoning, while open methods are testable but underpowered.

    \begin{principle}
        Static evaluation is insufficient for measuring general intelligence.
    \end{principle}
    \vspace{-0.2cm}
    \noindent ARC-AGI-3's interactive format exposes capabilities no static
    benchmark can assess, with an 8$\times$ larger human-AI gap than
    ARC-AGI-1. The full benchmark launches in March 2026.

    \vspace{+0.1cm}
    These insights inform Section~\ref{sec:empirical-analysis}.
\end{rqbox}


\section{Comparative Analysis}\label{sec:comparative-analysis}

We now compare approaches to surface the key trade-offs, recurring patterns,
and practical constraints that shape progress and inform future directions.
This comprehensive view reveals several patterns that inform our understanding
of the effectiveness of different approaches.

\subsection{Performance Overview by Paradigm}
\label{subsec:performance-overview}

Table~\ref{tab:arc-scores-by-category} presents aggregate performance statistics across
paradigms and representative individual systems.
First, the paradigm choice establishes a performance ceiling: pure transductive approaches cap
around 40\%, pure inductive methods reach 80\%, while hybrid approaches achieve the highest peaks
at 80-94\%. However, within-paradigm variance often exceeds between-paradigm differences. Inductive
approaches span from 2\% to 79.3\%, indicating that implementation quality matters as much as
architectural choice. The standard deviation within pure induction (27.4\%) exceeds the mean
difference between paradigms (approximately 15 percentage points), suggesting that how an approach
is implemented matters more than which paradigm is chosen.

\begin{table}[!t]
    \caption{Performance comparison across solution paradigms and representative systems
    on ARC-AGI-1 public evaluation set. Scores represent percentage of tasks solved exactly.
    Systems evaluated on $\geq$100 tasks provide more reliable estimates than those tested on
    smaller subsets. The paradigm choice establishes a performance ceiling, but within-paradigm
    variance often exceeds between-paradigm differences.}
    \label{tab:arc-scores-by-category}
    \centering
    \footnotesize
    \begin{tabular}{llr}
        \toprule
        \textbf{Paradigm} & \textbf{System} & \textbf{Score (\%)} \\
        \midrule
        \multicolumn{3}{c}{\textit{Pure Paradigm Approaches}} \\
        \hline
        Induction & Neural-guided synthesis~\cite{ouellette_out--distribution_2025} & 79.3 \\
        Induction & Abductive solver~\cite{lim_abductive_2024} & 66.0 \\
        Induction & ConceptSearch~\cite{singhal_conceptsearch_2025} & 50-58 \\
        Induction & Greenblatt GPT-4o~\cite{greenblatt_getting_2024} & 50.0 \\
        Induction & Graph constraints~\cite{xu_graphs_2022} & 35.6 \\
        Induction & MADIL~\cite{ferre_madil_2025} & 15.1 \\
        \midrule
        Test-Time Adapt. & Product of Experts~\cite{franzen_product_2025} & 71.6 \\
        Test-Time Adapt. & Deep learning (Cole)~\cite{cole_dont_2025} & 39.0 \\
        Test-Time Adapt. & Adaptive branching~\cite{inoue_wider_2025} & 12-16 \\
        \midrule
        Transduction & Atzeni et al.~\cite{atzeni_infusing_2023} & 80.0 \\
        Transduction & Hierarchical reasoning~\cite{wang_hierarchical_2025} & 40.3 \\
        Transduction & Video diffusion~\cite{acuaviva2025-from-g-to-g} & 16.8 \\
        Transduction & Neural CA~\cite{xu_neural_2025} & 13.4 \\
        \midrule
        \multicolumn{3}{c}{\textit{Hybrid Approaches}} \\
        \hline
        Induct.+TTA & Berman (NL programs)~\cite{berman2025-substack} & 79.6 \\
        Induct.+TTA & Pang (libraries)~\cite{pang2025-substack} & 77.1 \\
        Induct.+TTA & Huang ANPL~\cite{huang_anpl_2023} & 75.0 \\
        Induct.+TTA & ArcMemo~\cite{ho2025-arcmemo} & 56-59 \\
        Induct.+TTA & Self-improving LM~\cite{pourcel-2025-self-improving} & 52.0 \\
        Induct.+TTA & Test-time training~\cite{akyurek_surprising_2025} & 47.1 \\
        \bottomrule
    \end{tabular}
\end{table}

Second, cost efficiency varies dramatically and non-monotonically with performance (Table~\ref{tab:cost_perf_narrow}). While the current top-scoring system achieves the best headline performance at a moderate per-task cost, it represents a dramatic improvement over the prior frontier baseline from one year earlier~\cite{arcprize2025_leaderboard}.
Importantly, this $\sim390\times$ cost reduction appears to be driven largely by reduced parallelism (i.e., fewer samples per task) rather than fundamentally more efficient reasoning~\cite{arcprize2024_o3}. Below the frontier, systems such as Pang and NVARC illustrate that efficiency-oriented design can deliver competitive performance at practical costs~\cite{pang2025-substack,nvarc2025_kaggle}.
This cost-performance analysis reveals two distinct regimes: brute-force
test-time search approaches the performance ceiling but with diminishing returns, while
algorithmic approaches achieve practical performance within reasonable computational budgets.
\begin{table}[t]
    \caption{
        Cost-performance points showing non-monotonic efficiency;
        frontier gains can reflect reduced parallelism,
        while algorithmic designs reach practical costs.
    }
    \centering
    \footnotesize
    \setlength{\tabcolsep}{3pt}
    \begin{tabular}{lrrl}
        \hline
        \textbf{System} & \textbf{Score} & \textbf{\$/task} & \textbf{Note} \\
        \hline
        GPT-5.2 Pro & 90.5\% & 11.64 & $\sim$390$\times$ vs.\ o3; mainly fewer samples~\cite{arcprize2025_leaderboard} \\
        o3 & 87.5\% & 4,500 & High-cost frontier baseline~\cite{arcprize2024_o3} \\
        Gemini 3 Flash & 84.7\% & 0.17 & Best cost-efficiency ratio~\cite{arcprize2025_leaderboard} \\
        Pang & 77.1\% & 3.97 & Library-based program synthesis~\cite{pang2025-substack} \\
        NVARC & 24.03\% & 0.20 & Competition winner (ARC-AGI-2)~\cite{nvarc2025_kaggle} \\
        \hline
    \end{tabular}
    \label{tab:cost_perf_narrow}
\end{table}

Third, evaluation rigor critically affects reported performance. Systems evaluated comprehensively
on $\geq$100 tasks show mean performance approximately 27 percentage points lower than those evaluated
on $<$100 tasks (38.8\% vs 65.6\%). This 70\% relative inflation indicates that many reported high
scores come from cherry-picked subsets or insufficient statistical samples rather than robust
generalization. Competition-verified results on held-out test sets provide the most reliable
performance estimates, though fewer than 20\% of papers report such results. The few systems
reporting both public and private test performance reveal consistent gaps (typically 10-20
percentage points), suggesting overfitting to public evaluation characteristics.

\subsection{Success Factors and Failure Modes}
\label{subsec:success-factors}

Analysis of systems exceeding 70\% on ARC-AGI-1 reveals six empirically-observed
characteristics that correlate strongly with high performance
(Table~\ref{tab:success-factors-condensed}). These patterns are not strict
requirements but recur across all top-performing architectures regardless of
paradigm. Conversely, analysis of consistently failing approaches reveals
anti-patterns that predict poor performance regardless of implementation quality
(Table~\ref{tab:failure-modes-condensed}). Extended discussion of each factor
with supporting evidence is provided in the Supplementary Material.

\begin{table}[t]
\centering
\small
\setlength{\tabcolsep}{4pt}
\renewcommand{\arraystretch}{1.15}
\begin{tabular}{p{0.20\linewidth} p{0.48\linewidth} p{0.26\linewidth}}
\hline
\textbf{Success factor} & \textbf{Description} & \textbf{Evidence} \\
\hline
Guided search & Learned heuristics, neural nets, or LLMs prioritize promising candidates & Unguided search rarely exceeds 30\%; guided systems reach 72--80\% \\
Multiple hypotheses & Diverse candidates maintained; selected via demo consistency & Franzen's Product-of-Experts (71.6\%); Berman, Pang generate \(\ge\)10 candidates \\
Test-time adaptation & Fine-tuning, dynamic prompts, or adaptive search per task & Present in \emph{every} \(>70\%\) system; \(\sim\)20--30 pt gain over frozen baselines~\cite{akyurek_surprising_2025} \\
Hybrid reasoning & Neural perception + symbolic verification and composition & Pure transduction caps \(\sim\)40\%; pure induction \(\le\)79\%; hybrids reach 80--94\% \\
Robust perception & Multiple candidate representations (objects, patterns, regions) & 20--30 pt advantage over single-representation systems \\
Compositional primitives & Reusable operations composed for novel tasks & Enables systematic generalization; limited to 1--2 steps in current systems \\
\hline
\end{tabular}
\caption{Success factors shared by systems exceeding 70\% on ARC-AGI-1.}
\label{tab:success-factors-condensed}
\end{table}

\begin{table}[t]
\centering
\small
\setlength{\tabcolsep}{4pt}
\renewcommand{\arraystretch}{1.15}
\begin{tabular}{p{0.20\linewidth} p{0.48\linewidth} p{0.26\linewidth}}
\hline
\textbf{Anti-pattern} & \textbf{Description} & \textbf{Impact} \\
\hline
Pure pattern matching & Learns correlations, not causal/compositional rules & \(\sim\)40\% ceiling; collapses on harder compositions \\
Unguided exhaustive search & Enumerates programs without strong heuristics & Exponential blowup; usually \(\le\)30\% \\
Static representation & Locks into one abstraction (objects/regions/etc.) & Brittle failures; \(\sim\)25-pt deficit \\
No test-time adaptation & Fixed policy; ignores task demos as learning signal & \(\sim\)30--40\% plateau; \(\sim\)20--30 pts behind adaptive \\
Monolithic architecture & No modular hybrid of perception + symbolic ops & \(\sim\)20--40 pts worse than hybrids \\
Frontier inefficiency & High score via expensive compute; hardness cliff persists & Costly per task; far from human efficiency \\
\hline
\end{tabular}
\caption{Common anti-patterns and their observed impacts.}
\label{tab:failure-modes-condensed}
\end{table}

\subsection{The Performance Cliff: ARC-AGI-2 and ARC-AGI-3}
\label{subsec:performance-cliff}

%
\begin{figure}[!t]
    \centering
    \includegraphics[width=0.85\textwidth]{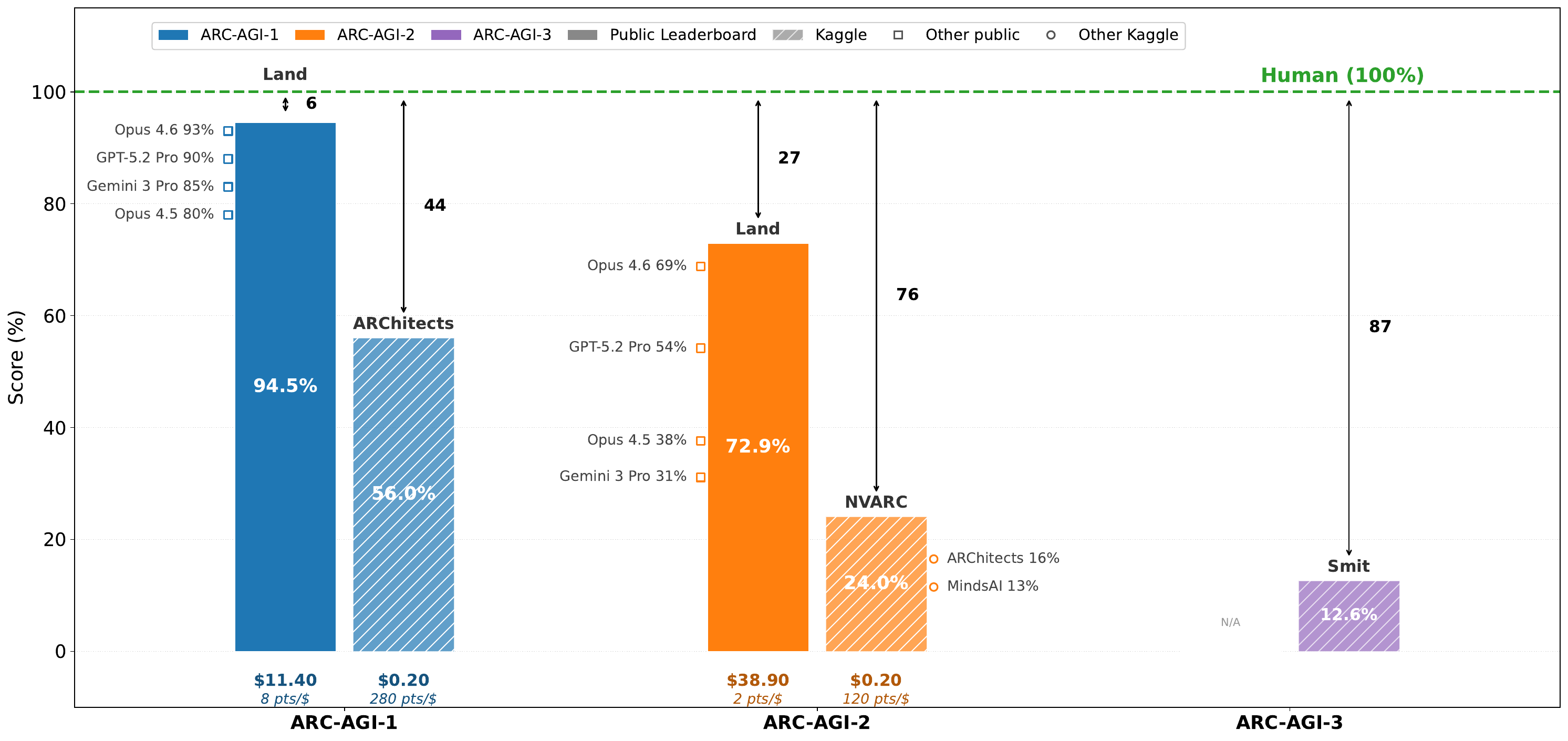}
    \caption{
        ARC-AGI performance across benchmark versions. Public leaderboard
        (solid bars) allows unconstrained compute and API access; Kaggle competition
        (hatched bars) is constrained to \$50 compute budget with no internet.
        Arrows show gap to human baseline. The human baseline of 100\% represents
        task-level solvability (every task solved by at least one person); average
        individual accuracy is 76\% on ARC-AGI-1 and 60\% on ARC-AGI-2.
        Key findings: (1)~Land's cross-model ensemble~\cite{land2025_arc_solver}
        sets public SOTA at 94.5\% (ARC-AGI-1) and 72.9\% (ARC-AGI-2); (2)~Opus~4.6
        nearly matches Land at 93.0\% and 68.8\% respectively, at a fraction of
        the cost (\$1.88 and \$3.64/task vs.\ \$11.40 and \$38.90/task); (3)~Public
        scores exceed Kaggle by 30--70 percentage points, reflecting unconstrained
        vs.\ constrained compute regimes; (4)~The performance cliff persists across
        all systems: even the best public entry drops 23\% from ARC-AGI-1 to ARC-AGI-2;
        (5)~Kaggle winners achieve better cost-efficiency: NVARC scores 24\% at
        \$0.20/task (120~pts/\$) vs.\ Land's 72.9\% at \$38.90/task (2~pts/\$),
        a 60$\times$ efficiency gap.
    }
    \label{fig:performance-cliff-2}
\end{figure}


Performance on ARC-AGI-2 and ARC-AGI-3 exposes three critical limitations: (1) compositional
depth beyond 2-3 steps causes exponential search space growth that overwhelms current methods,
(2) context-dependent rule application remains poorly handled by all paradigms, and
(3) interactive exploration capabilities are largely absent from current architectures.

Of the total numbers of papers retained for evaluations (detailed in Section~\ref{subsec:evaluation-reliability}),
only 15 report ARC-AGI-2 results and 3 report ARC-AGI-3 results, reflecting these benchmarks'
recency and difficulty. The available data reveals consistent patterns across all paradigms.

\subsubsection{The ARC-AGI-2 Degradation Pattern}

ARC-AGI-2 performance shows 2.5--3$\times$ degradation compared to ARC-AGI-1, regardless of approach
type or implementation quality. The ARC Prize 2025 competition~\cite{arcprize2025_results}
confirmed this pattern: on the private Kaggle test set, NVARC achieved 24.03\%, The ARChitects
16.53\%, and MindsAI 12.64\%. The public leaderboard~\cite{arcprize2025_leaderboard} showed
commercial systems achieving higher scores (Poetiq 54\%, Opus 4.5 37.6\%), yet even these
represent substantial degradation from ARC-AGI-1 performance levels. This consistent decline
reveals fundamental limitations in compositional reasoning rather than overfitting, as these
systems were competition-verified on held-out test sets. Section~\ref{subsec:arc-prize-2025-winners}
summarizes the 2025 winning approaches (detailed architectures in the Supplementary Material).

ARC-AGI-2's primary difference from ARC-AGI-1 lies in increased compositional complexity:
average transformation depth increases from 1.3 to 2.7 steps, while maintaining the same core
knowledge priors and task format. Tasks still involve grids with simple geometric and color
transformations, still require the same foundational cognitive abilities, and still provide 3-5
demonstration examples. The only systematic difference is that solutions require composing more
primitive operations. This controlled variation isolates compositional generalization as the
limiting factor.

Wang's hierarchical neural approach shows the sharpest decline (40.3\% $\rightarrow$ 5\%), 
suggesting that pure neural methods fail earliest as compositional demands increase~\cite{wang_hierarchical_2025}. 
The 87\% relative drop indicates that pattern matching approaches, regardless 
of architectural sophistication, cannot handle even modestly deeper compositions, 
a finding consistent with the broader limitations of pure transductive methods.

\begin{figure}[!t]
    \centering
    \includegraphics[width=0.85\linewidth]{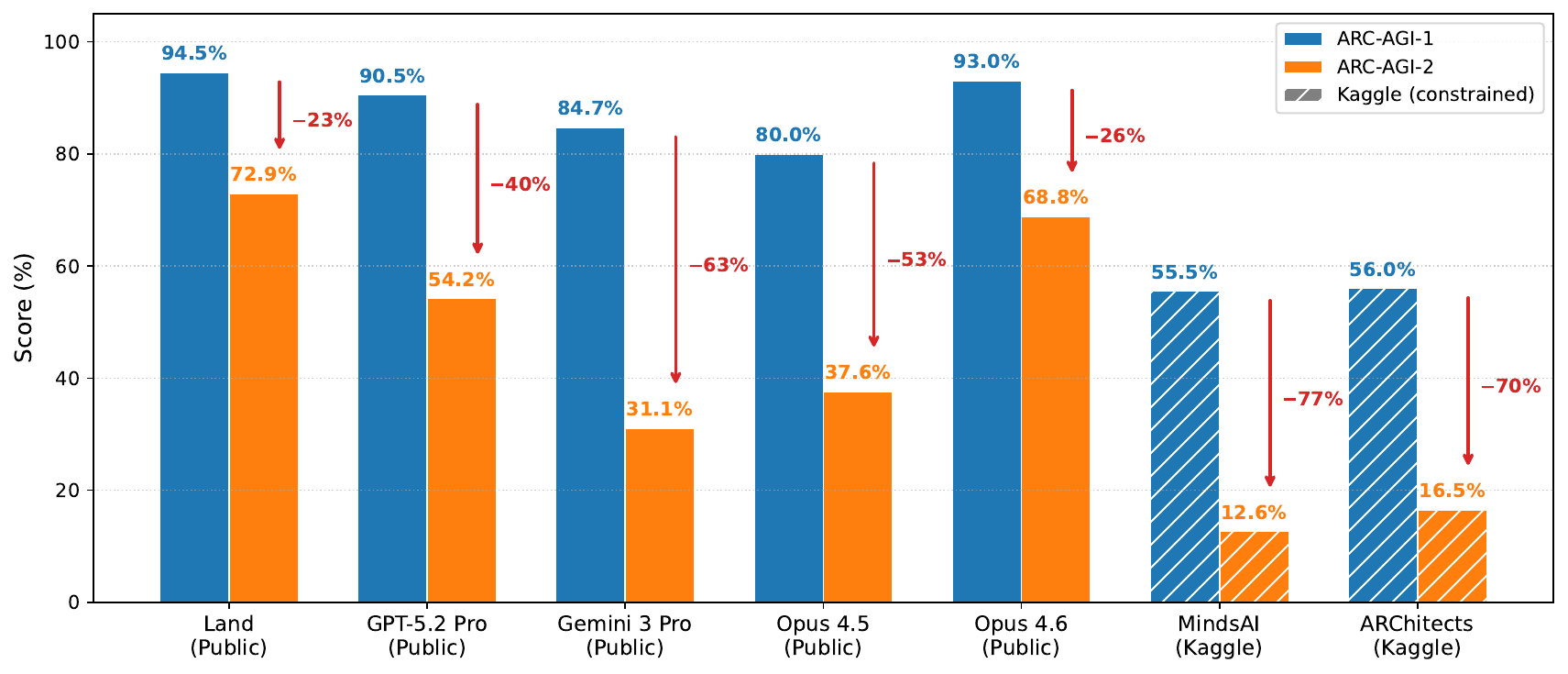}
    \caption{
        Cross-generation performance cliff: seven systems evaluated on both
        ARC-AGI-1 and ARC-AGI-2. Land's cross-model
        ensemble~\cite{land2025_arc_solver} achieves the highest scores on
        both benchmarks but still drops 23\%. Notably, Opus~4.6 shows the
        smallest single-model drop ($-$26\%, from 93.0\% to 68.8\%),
        compared to its predecessor Opus~4.5 ($-$53\%) and GPT-5.2~Pro
        ($-$40\%). Other single frontier models (solid bars) drop 40--63\%,
        while Kaggle competition winners (hatched bars) drop 70--77\%.
        The human baseline of 100\% represents task-level solvability
        (every task solved by at least one person); average individual
        accuracy is 76\% on ARC-AGI-1 and 60\% on
        ARC-AGI-2~\cite{chollet_arc-agi-2_2025}.
    }
    \label{fig:scaling-question}
\end{figure}

Figure~\ref{fig:scaling-question} visualizes this degradation for six systems
spanning a cross-model ensemble (\cite{land2025_arc_solver}),
single frontier models (GPT-5.2 Pro,
Gemini 3 Pro, Opus 4.5), and resource-constrained Kaggle competition winners
(MindsAI~\cite{mindsai2025_kaggle}, ARChitects~\cite{the_architects_2025_techical_report}).
The relative drops range from 23\% (Land) to 77\%
(MindsAI), yet the pattern is universal: no system, regardless of scale,
cost regime, or architectural paradigm, maintains its ARC-AGI-1 performance
on ARC-AGI-2.

\subsubsection{Why Current Composition Strategies Fail}

The root cause is that current systems treat composition as sequential concatenation,
searching over all ordered sequences of operations. Search spaces grow exponentially
with depth ($O(n^d)$ for $n$ operations and depth $d$): a 2-step transformation
requires searching pairs, a 3-step transformation requires triples, quickly
exceeding computational budgets. Human reasoning, by contrast, employs hierarchical
decomposition that scales sub-exponentially by reducing the effective search space at
each level. No current architecture implements explicit mechanisms for such hierarchical
reasoning, explaining the consistent performance cliff across all
paradigms (Section~\ref{subsec:remaining-challenges}).

\subsubsection{ARC-AGI-3 and Interactive Intelligence}
ARC-AGI-3's shift to interactive environments
exposes capability gaps beyond compositional reasoning. The preview benchmark's top
system achieves only 12.58\% action efficiency~\cite{arcagi3_learning}, the lowest
performance across all ARC versions, revealing that static-to-interactive transfer
demands capabilities current architectures largely lack: hypothesis-driven exploration,
persistent memory across interaction sequences, goal inference from sparse feedback,
and cross-game transfer. Where ARC-AGI-2 tests whether systems can compose known
primitives, ARC-AGI-3 tests whether they can \emph{discover} those primitives
autonomously, compounding the compositional challenge with an exploration
challenge.

\begin{rqbox}{Summary of Comparative Analysis.}
    The ARC-AGI benchmark's three-generation evolution provides a diagnostic lens
    on these capabilities. Performance cliffs persist across versions but are
    scale-dependent: frontier models with massive compute narrow the gap
    substantially, while constrained systems show steep degradation, suggesting
    that brute-force coverage partially compensates for missing compositional
    mechanisms. Achieving robust, efficient performance will require architectural
    innovations addressing compositional reasoning, hierarchical decomposition,
    and adaptive abstraction. ARC-AGI-3's interactive paradigm further raises the
    bar by demanding exploration and world model induction, capabilities that no
    current architecture demonstrates at human-level efficiency.
\end{rqbox}

\section{Empirical Analysis}\label{sec:empirical-analysis}

This section presents our empirical analysis of approaches to ARC-AGI, examining
the temporal evolution of performance from 2019 through 2025, the dramatic performance
variations across benchmark generations, and the economic and methodological
dimensions of current research. Drawing on quantitative data from 80 surveyed papers,
we trace the field's progression from initial failures to recent breakthroughs,
identify critical inflection points, and assess the reliability of reported results.
We apply exclusion criteria selectively for visualizations to avoid misleading
interpretations of empirical performance. In particular, we exclude from score-based
plots: (1) studies not scalable to the full ARC-AGI dataset, (2) results reported
on fewer than 100 tasks, and (3) diverse-inference settings that can artificially
inflate accuracy (e.g., near-perfect scores using multiple agents). After applying
these filters, 59 papers remain for quantitative performance comparisons
(see Supplementary Material, Section~2, for the complete selection workflow).
Our analysis reveals both the remarkable progress achieved on ARC-AGI-1 and the
persistent fundamental challenges exposed by compositionally more complex variants.

\subsection{Temporal Evolution of Performance (2019-2025)}
\label{subsec:temporal-evolution}

The history of ARC-AGI research exhibits a striking pattern: five years of
incremental progress followed by a dramatic six-month breakthrough period,
culminating in performance that approaches but cannot surpass human baselines
on the original benchmark. This temporal trajectory, illustrated in
Figure~\ref{fig:scores_over_time}, reveals how concentrated research effort
catalyzed by competitive incentives can accelerate progress, while also
exposing fundamental limitations that resist such acceleration.

From 2019 through 2023, research efforts produced modest gains despite
substantial methodological diversity. Maximum accuracy across 25 documented
approaches remained below 20\%, with symbolic program synthesis methods
achieving the highest performance through domain-specific languages and
brute-force search. Neural approaches during this period consistently
underperformed, rarely exceeding 5\% accuracy, leading to widespread
skepticism about deep learning's applicability to abstract reasoning tasks.
The field's primary contributions during this era came not in performance
gains but in methodological exploration: establishing baseline approaches,
identifying core challenges, and developing the conceptual frameworks that
would later enable breakthroughs.

The emergence of large language models in early 2024 catalyzed a qualitative
shift in approach philosophies. Researchers began exploring LLM-guided program
synthesis, leveraging models' code generation capabilities to navigate vast
program spaces more intelligently than exhaustive search. Systems like
ConceptSearch achieved 58\% accuracy by using natural language descriptions
to guide exploration, while Greenblatt's GPT-4o-based approach reached 50\%
through carefully engineered prompting strategies. These results demonstrated
that pre-trained models, despite never encountering ARC tasks during training,
contained inductive biases useful for compositional reasoning, a finding that
challenged prevailing assumptions about the necessity of task-specific
architectures.

\definecolor{starblue}{HTML}{3498DB}

\begin{figure}[!t]
  \centering
  \begin{tikzpicture}
    \node[anchor=south west, inner sep=0] (img)
      {\includegraphics[width=0.85\textwidth, trim=20 30 10 20,clip]{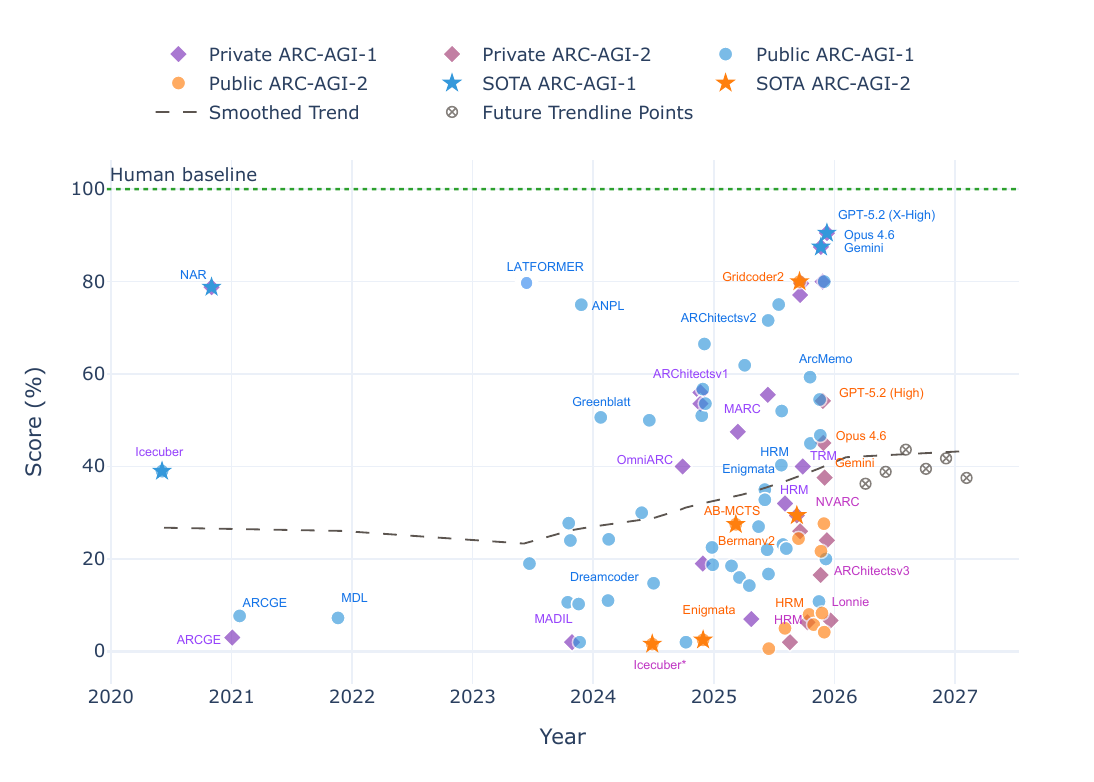}};
    \begin{scope}[x={(img.south east)}, y={(img.north west)}]
      \draw[dashed, line width=.75pt, orange] (0.59,0.10) -- (0.59,0.82);
      \draw[dashed, line width=.75pt, cyan!80!black] (0.065,0.10) -- (0.065,0.82);

      \node[
        star,
        star points=5,
        star point ratio=0.35,
        fill=starblue,
        draw=none,
        minimum size=0.02cm,
        rotate=180,
      ] at (0.76,0.74) {};
    \end{scope}
  \end{tikzpicture}
    \caption{
        Temporal evolution of ARC-AGI performance (2019--2025). Blue: ARC-AGI-1;
        Amber: ARC-AGI-2. The 2024 phase transition shows performance improving more
        in six months than the previous five years. The consistent 2.5--3$\times$ degradation
        from ARC-AGI-1 to ARC-AGI-2 across all approaches indicates fundamental compositional limitations. 
    }
    \label{fig:scores_over_time}
\end{figure}
The ARC Prize 2024 competition, launched in June with a \$1 million prize pool,
triggered an unprecedented research acceleration. Over 1,430 teams participated,
exploring diverse strategies unified by a common recognition: test-time
adaptation matters more than training-time scale. The competition's top
performer, MindsAI, achieved 55.5\% through test-time fine-tuning
combined with inference-time augmentation. Notably, all teams in the top 50
exceeded 40\% accuracy and employed some form of test-time adaptation,
validating this technique as essential rather than auxiliary.

The ARC Prize 2025 competition~\cite{arcprize2025_results} built on this momentum with
even greater participation: 1,455 teams submitted 15,154 entries, and 90 papers were
submitted (up from 47 in 2024). On the harder ARC-AGI-2 benchmark, the private Kaggle
competition saw NVARC achieve 24.03\% (1st place, \$25k), The ARChitects 16.53\% (2nd,
\$10k), and MindsAI 12.64\% (3rd, \$5k). The public leaderboard~\cite{arcprize2025_leaderboard}
showed commercial systems achieving higher scores: Poetiq's Gemini 3 Pro refinement reached
54\% at \$30.57/task, Google's Gemini 3 Deep Think achieved 45.1\% at \$77.16/task, and
Anthropic's Opus 4.5 reached 37.6\% at \$2.20/task. These competitions demonstrated that
focused incentives combined with open dataset access could accelerate progress dramatically,
though the performance gap between public and private test sets highlights ongoing
challenges with robust generalization.

December 2024 marked another inflection point with OpenAI's announcement of o3,
achieving 87.5\% on ARC-AGI-1 through massive test-time search. This result
required 1,024 parallel samples per task at approximately \$4,500 per task,
establishing a new performance ceiling while highlighting severe efficiency
challenges~\cite{chollet_arc_2025}. In December 2025, OpenAI's GPT-5.2 Pro achieved
90.5\% at just \$11.64/task, a 390$\times$ efficiency improvement~\cite{arcprize2025_leaderboard}.
However, this cost reduction likely stems largely from reduced parallelism rather
than fundamentally more efficient reasoning: o3's ``low efficiency'' mode used
1,024 samples and 5.7B tokens per 100 tasks, while ``high efficiency'' used only
6 samples at 33.5M tokens (achieving 75.7\% vs.\ 87.5\%). The 390$\times$ improvement
thus reflects engineering optimization (fewer parallel runs at lower API costs) rather
than algorithmic compression toward human-like efficiency. In parallel, program
synthesis approaches demonstrated that high performance could be achieved efficiently:
Berman's evolutionary system reached 79.6\% at \$8.42 per task through natural
language instruction evolution, while Pang's library-based method achieved 77.1\%
at \$3.97 per task.

\begin{figure}[!t]
    \centering
    \includegraphics[width=\textwidth]{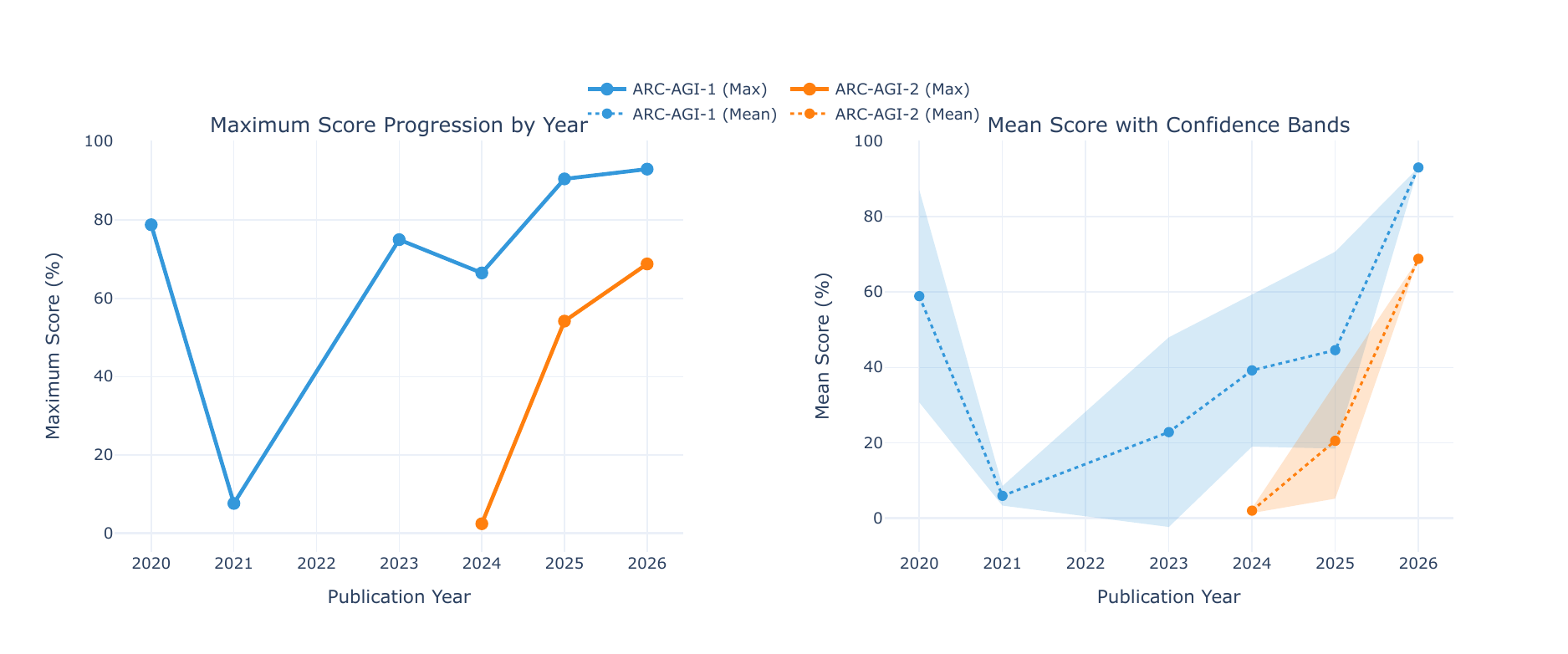}
    \caption{
            Yearly progress of ARC-AGI scores from 2020 to 2026. Left:
            maximum score progression. ARC-AGI-1 scores rose from 78.8\% (2020)
            through a plateau in 2023--2024 before surging to 93.0\% in 2026
            (Opus~4.6); ARC-AGI-2, introduced in 2024, jumped from 2.5\% to 68.8\%
            in under two years. Right: mean scores with $\pm 1\sigma$ confidence
            bands. The widening bands in 2024--2025 (ARC-AGI-1 mean 39--45\%,
            $\sigma \approx 20$--26) reflect growing methodological diversity as
            the field expanded from 2--3 papers per year to over 40. The persistent
            gap between maximum and mean scores indicates that top-performing
            techniques (large-scale inference, test-time training) have not yet
            diffused broadly across the research community.
    }
    \label{fig:yearly_progress}
\end{figure}

The January 2025 release of ARC-AGI-2 initially halted this triumphant narrative.
Performance on the new benchmark dropped significantly: Berman achieved 29.4\%,
Pang reached 26.0\%, and other systems struggled to exceed 25\%. By December 2025,
GPT-5.2 Pro achieved 54.2\% on ARC-AGI-2 at \$15.72/task~\cite{arcprize2025_leaderboard},
nearly matching the best refinement-based approaches. Yet even this represents a
substantial drop from 90.5\% on ARC-AGI-1, confirming that the 2.5-3$\times$ degradation
pattern persists despite the efficiency revolution. The consistency of this
performance cliff across all paradigms (program synthesis, neuro-symbolic hybrids,
and neural approaches alike) indicates a shared fundamental limitation in
compositional generalization rather than paradigm-specific weaknesses.

The July 2025 preview of ARC-AGI-3 exposed additional capability gaps by shifting
from static puzzles to interactive mini-games requiring exploration and goal
inference. The highest reported performance, 12.58\% by StochasticGoose, represented
the lowest achievement across all ARC versions, despite tasks remaining within
the scope of human core knowledge priors. This results revealed that
capabilities developed for passive pattern recognition do not transfer to active
learning contexts, suggesting that current architectures miss essential aspects
of intelligence related to reasoning and interaction.

\begin{figure}[!t]
\centering
\includegraphics[trim={2cm 1cm 2cm 1cm}, width=0.85\textwidth]{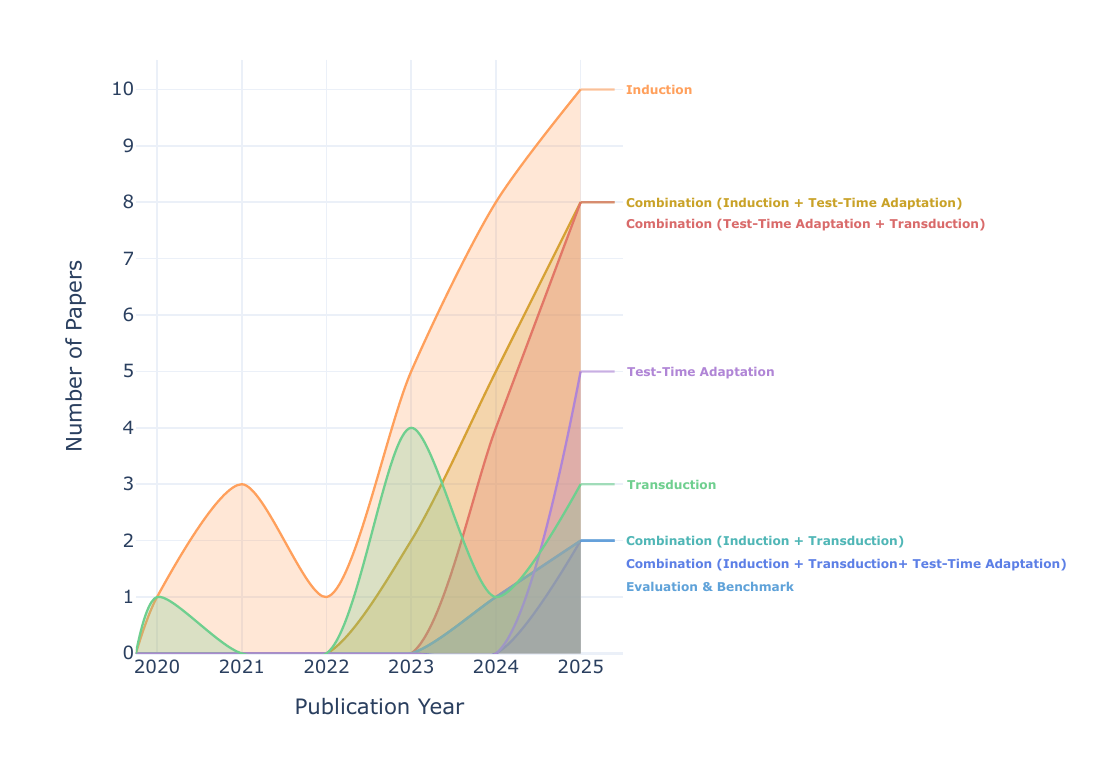}
\caption{
    Publication trends by research category. The 2024--2025 surge reflects the
    ARC Prize competition's catalytic effect. Hybrid approaches dominate,
    reflecting recognition that neither pure neural nor symbolic methods suffice.
    Pure neural approaches declined after demonstrating consistent ~20\% ceilings.
}
\label{fig:papers_by_category}
\end{figure}

Figure~\ref{fig:yearly_progress} shows the widening variance in 2024--2025 reflecting
methodological diversity, while Figure~\ref{fig:papers_by_category} quantifies
publication trends. The dominance of hybrid approaches reflects collective
recognition that combining neural perception with symbolic reasoning provides
capabilities beyond either pure paradigm. The field remains in an exploratory
phase rather than having converged on superior approaches.

Examining performance across ARC-AGI versions provides critical diagnostic
information about the nature and limitations of current reasoning capabilities.
The three benchmark generations employ identical core knowledge priors and
presentation formats, varying primarily in compositional complexity (ARC-AGI-2)
and interaction requirements (ARC-AGI-3). This controlled variation isolates
specific capability dimensions, revealing which aspects of abstract reasoning
current systems have mastered and which remain fundamentally challenging.

The comprehensive performance table (Supplementary Material, Table~1) presents 
comprehensive performance data across paradigms, years, and benchmark versions. 
Several patterns emerge that illuminate the current state of the field. First, 
paradigm choice establishes a performance ceiling but does not guarantee success: 
pure inductive approaches span from 2\% to 79.3\%, indicating that implementation 
quality often matters more than architectural category. Second, hybrid approaches 
demonstrate more consistent performance, clustering in the 50-75\% range on ARC-AGI-1 while
avoiding the catastrophic failures common in pure approaches. Third, the
scarcity of ARC-AGI-2 results (only 15 of 80 papers report performance) reflects
both the benchmark's recency and its difficulty; researchers may be reluctant to
publish results on benchmarks where their methods perform poorly.

The comprehensive performance data for all 59 papers is provided in the
Supplementary Material (Table~1).

The performance degradation from ARC-AGI-1 to ARC-AGI-2 exhibits remarkable
consistency across diverse approaches. Systems employing program synthesis
(Berman: 79.6\% $\rightarrow$ 29.4\%), neural methods (Wang: 40.3\% $\rightarrow$ 5\%), and hybrid
architectures (Franzen: 60.5\% $\rightarrow$ 2.5\%) all show 2.5-3$\times$ or greater performance
drops. This cross-paradigm consistency indicates shared fundamental limitations
rather than paradigm-specific weaknesses addressable through architectural
refinements. The magnitude of degradation correlates with the compositional
depth increase in ARC-AGI-2, suggesting that current systems' compositional
reasoning capabilities remain shallow despite sophisticated engineering.

\begin{figure}[!t]
    \centering
    \includegraphics[width=0.85\textwidth, trim= 20 20 40 70,clip]{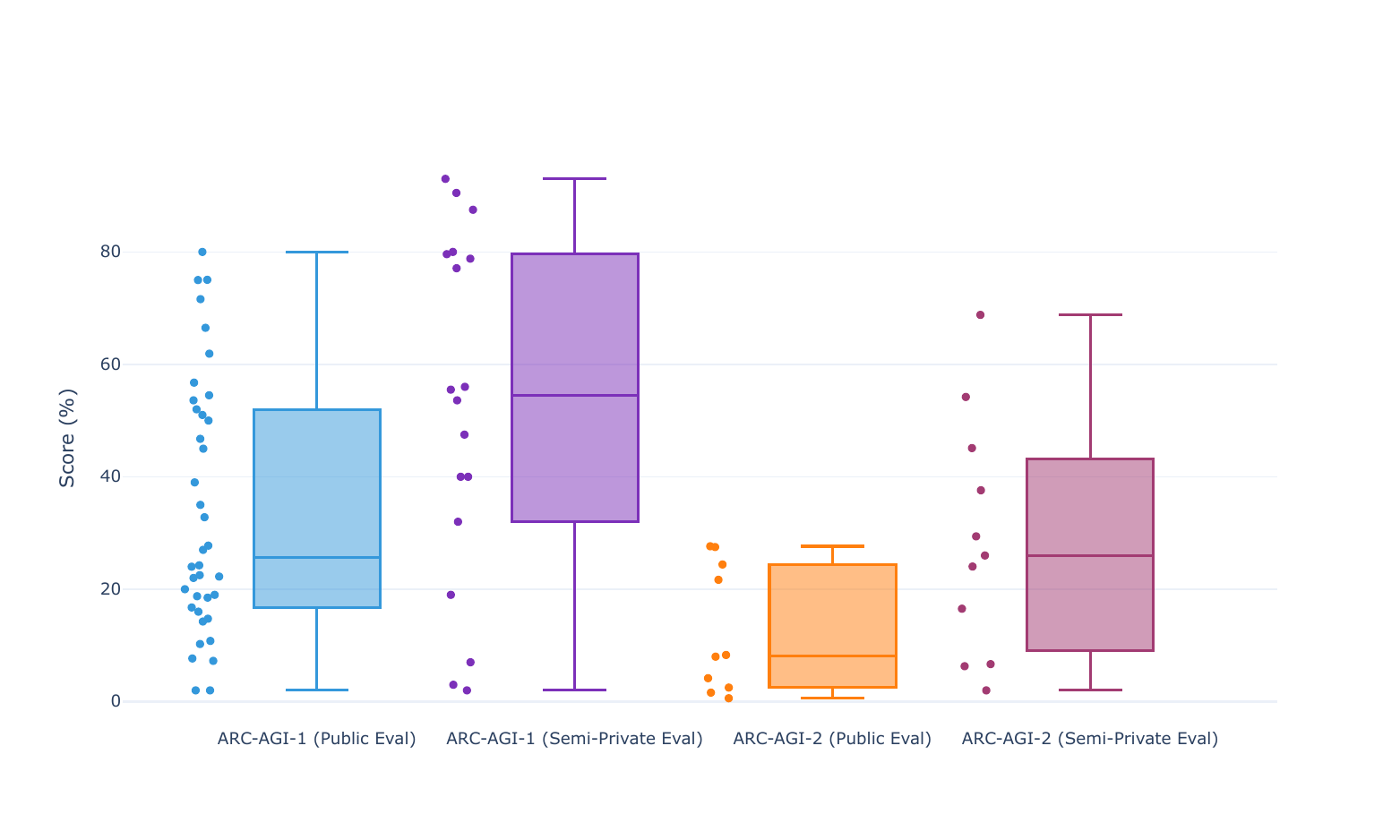}
    \caption{
        Score distributions across ARC-AGI benchmark versions and
        evaluation sets. Public eval scores (self-reported on the public
        evaluation set) cover 38 systems on ARC-AGI-1 (median~25.6\%,
        mean~34.1\%) and 10 on ARC-AGI-2 (median~8.2\%, mean~12.6\%).
        Semi-private eval scores (official leaderboard)
        appear higher because the two groups represent \emph{different
        populations}: frontier models (Opus~4.6, GPT-5.2~Pro, Gemini~3)
        submitted exclusively to the semi-private eval and never published
        public eval scores, skewing that distribution upward (18 systems on
        ARC-AGI-1, median~54.5\%; 11 on ARC-AGI-2, median~26.0\%). Among
        the 8 systems reporting \emph{both} scores on ARC-AGI-1,
        semi-private results are 5--38 percentage points \emph{lower} than
        public, consistent with the harder held-out set.
    }
    \label{fig:score_distributions}
\end{figure}

Analyzing performance distributions provides additional insights into systematic
patterns. Figure~\ref{fig:score_distributions} shows that ARC-AGI-1 public scores
exhibit high variance (standard deviation: 24.7\%) with a long tail of low-performing
approaches below 20\%. The median performance of approximately 40\% indicates that
achieving human-competitive results remains challenging for typical approaches,
with only the top quartile exceeding 60\%. ARC-AGI-2 scores cluster consistently
below 30\% with minimal variance, indicating that increased compositional complexity
creates a performance ceiling that few current methods can overcome. The gap between
public and private scores, while sparsely documented (only 3 systems report both),
suggests potential overfitting to public test characteristics, though limited data
prevents definitive conclusions.

Statistical analysis reveals paradigm-specific patterns. Pure inductive approaches
exhibit the highest variance ($\sigma$=27.4\%, range: 0--79.3\%), highly sensitive to
implementation details like search strategy and guidance mechanisms. Pure
transductive approaches show lower variance ($\sigma$=12.8\%) but ceiling around 40\%.
Hybrid approaches achieve both the highest median (56\%) and most consistent
results, though even the best show severe ARC-AGI-2 degradation.
The temporal dimension reveals field-level learning curves: steady improvement
from 2019--2023 (~5\%/year), dramatic 2024 jump (35\% gain), then 2025 plateau,
suggesting diminishing returns as performance approaches current paradigm limits.

\subsection{Cost-Performance Frontiers}
\label{subsec:cost-performance-frontiers}

\begin{figure}[!t]
\centering
\includegraphics[width=0.8\textwidth, trim= 20 30 10 30, clip]{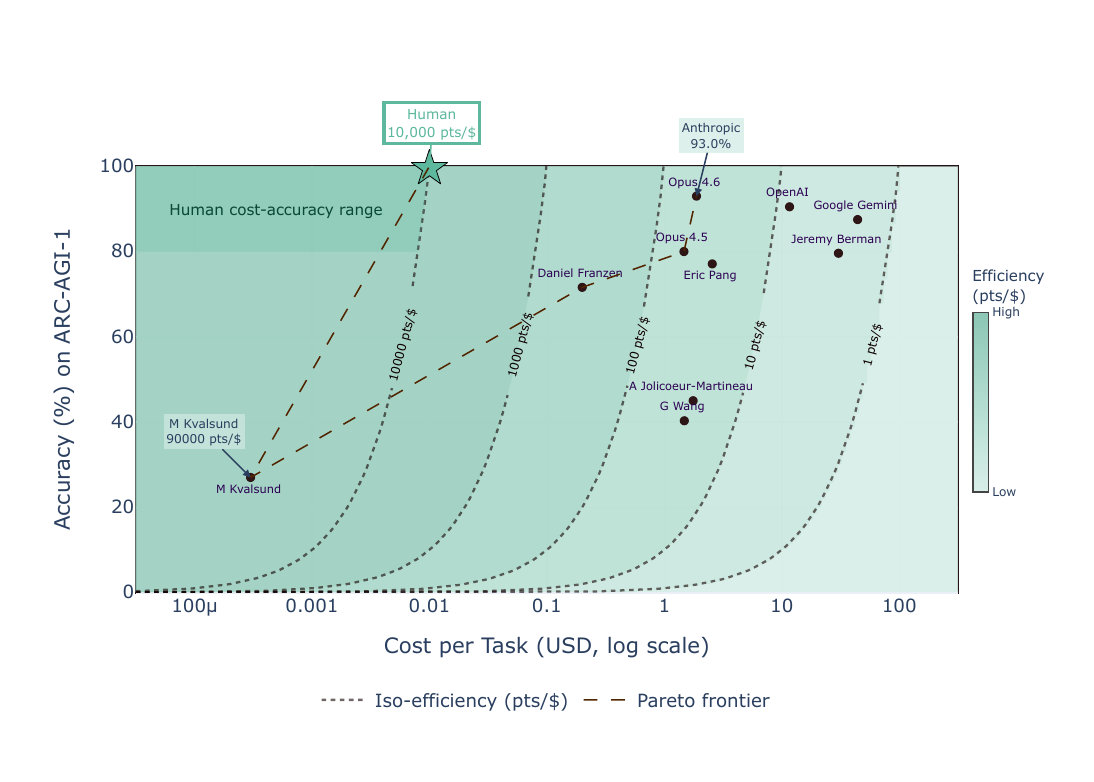}
\caption{
        Cost per task versus accuracy on ARC-AGI-1 for the 10
        systems (12\% of surveyed papers) reporting cost data. Dotted
        lines are iso-efficiency curves (constant pts/\$). Costs span
        five orders of magnitude (\$0.0003--\$44/task) with efficiencies
        ranging from 2 to 90,000~pts/\$. The Pareto frontier (dashed)
        shows steep diminishing returns above \$2/task. The human
        baseline (star) occupies the high-efficiency region that no AI
        system yet approaches.
}
\label{fig:cost_vs_performance}
\end{figure}

Economic analysis of ARC-AGI approaches reveals critical trade-offs between
performance and computational resources, with implications for both practical
deployment and theoretical understanding of intelligence. The cost-performance
landscape, visualized in Figure~\ref{fig:cost_vs_performance}, exhibits three
distinct regimes characterized by different scaling behaviors and economic
viability. Understanding these regimes illuminates not only practical considerations
for system design but also fundamental questions about the nature of reasoning:
whether performance gains come from algorithmic insight or merely from increased
computational expenditure.

The low-cost regime (<\$1/task) achieves approximately 40\% through efficient methods like
small fine-tuned models or constraint-based synthesis. Notably, Gemini 3 Flash Preview
achieves 84.7\% on ARC-AGI-1 at just \$0.17/task and 33.6\% on ARC-AGI-2 at
\$0.23/task, establishing a new cost-efficiency Pareto frontier that rivals
competition winners at substantially lower cost. The practical regime
(\$1--50/task) includes GPT-5.2 Pro at 54.2\%/\$15.72 (ARC-AGI-2) and
90.5\%/\$11.64 (ARC-AGI-1), Opus 4.5 at 37.6\%/\$2.20, and Poetiq at 54\%/\$30.57.
Human performance costs approximately \$0.30--0.60/task (1--2 minutes at median wages),
so AI systems require 10$\times$--100$\times$ human cognitive costs for comparable performance.

The frontier regime (>\$100/task) previously exhibited severe diminishing returns:
OpenAI's o3 achieved 75.7\% at \$26/task (6 samples) versus 87.5\% at \$4,560/task
(1,024 samples), a 175$\times$ cost increase for 12 percentage points. The 390$\times$ efficiency
improvement from o3 to GPT-5.2 Pro (90.5\% at \$11.64/task) appears dramatic but
likely reflects reduced parallelism rather than algorithmic breakthroughs. The
cost-performance curve evolves through engineering optimization, yet the fundamental
scaling challenge persists: achieving the final percentage points toward human-level
performance still requires disproportionate resources.

Scaling analysis shows performance follows $\text{Accuracy} = \alpha + \beta \log(\text{Cost})$
with $\beta \approx 0.15$: each 10$\times$ cost increase yields ~15 percentage points,
making scaling to human-level (100\%) economically prohibitive. Algorithmic
improvements provide far more efficient paths: Pang's library approach (\$3.97,
77.1\%) outperforms systems costing 10$\times$ more, and Berman's shift from code to
natural language evolution achieved 26 percentage point gains with minimal cost
increase.

Critical transparency limitations persist: only 9 of 80 papers (11\%) report both
cost and performance, potentially biasing the field toward expensive approaches.
The efficiency gap between human (~20W, 1--2 minutes, ~2,400J/task) and AI
(megawatt-scale, 6--7 orders of magnitude more energy) reveals that current
approaches achieve performance through exhaustive search rather than discovering
minimal, transferable abstractions.

\begin{figure}[!t]
    \centering
    \includegraphics[width=0.85\linewidth]{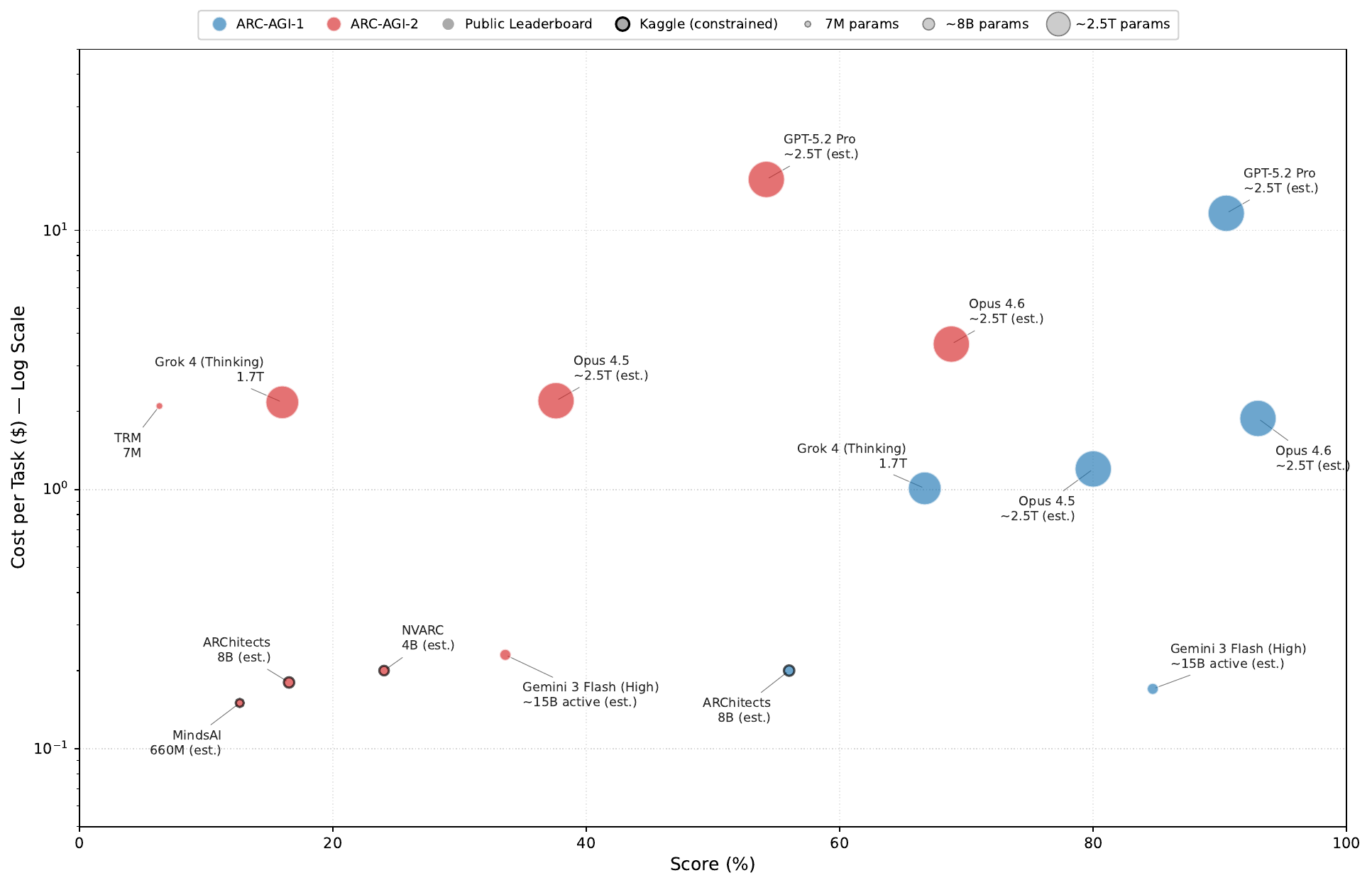}
    \caption{
        Score vs.\ cost per task vs.\ estimated model scale across
        ARC-AGI-1 and ARC-AGI-2. Bubble area is proportional to
        parameter count (log-scaled) and black edges denote
        Kaggle-constrained entries. Parameter counts marked
        \textit{(est.)} are community estimates; undisclosed
        architectural choices make cross-family comparisons approximate.
        On ARC-AGI-1, trillion-scale models vary widely in both score
        and cost: Opus~4.6 reaches 93.0\% at \$1.88/task, GPT-5.2~Pro
        scores 90.5\% at \$11.64/task, and Grok~4 scores 66.7\% at
        \$1.01/task.
        Kaggle-constrained entries cluster at \$0.15--\$0.20/task with
        660M--8B parameters, while TRM reaches 6.3\% on ARC-AGI-2 with
        just 7M parameters.
        This aligns with Chollet's thesis underlying
        ARC-AGI~\cite{chollet_measure_2019}: ``solely measuring skill
        at any given task falls short of measuring intelligence, because
        skill is heavily modulated by prior knowledge and experience.''
    }
    \label{fig:efficiency-bubble}
\end{figure}

Adding model scale as a third dimension (Figure~\ref{fig:efficiency-bubble})
further sharpens this picture: the relationship between parameter count, cost,
and score reveals that larger models do not proportionally outperform smaller
ones on abstract reasoning tasks.

\subsection{Evaluation Practices and Reliability}
\label{subsec:evaluation-reliability}

Rigorous evaluation practices prove essential for accurately assessing progress
toward artificial general intelligence, yet significant variance in evaluation
standards across published work complicates cross-study comparisons and may lead
to misleading performance claims. Our meta-analysis of evaluation methodologies
reveals systematic biases that inflate reported performance, limited transparency
in reporting practices that prevents reproducibility, and substantial gaps in
generalization assessment that obscure true capability levels.

Standard ARC-AGI evaluation protocols specify: exact match scoring (no partial
credit), maximum two attempts per task, held-out test sets, and separate public/private
reporting. Comprehensive reporting should include cost, inference time, and
training data requirements.

However, adherence varies dramatically. Systems evaluated on $\geq$100 tasks report
mean performance of 38.8\%, while those on <100 tasks report 65.6\%, a 26.8
percentage point inflation (70\% relative overestimation) arising from statistical
variance, task selection bias, and reduced power to detect overfitting.

\begin{figure}[!t]
\centering
\includegraphics[width=0.90\textwidth, trim=20 30 30 30 ,clip]{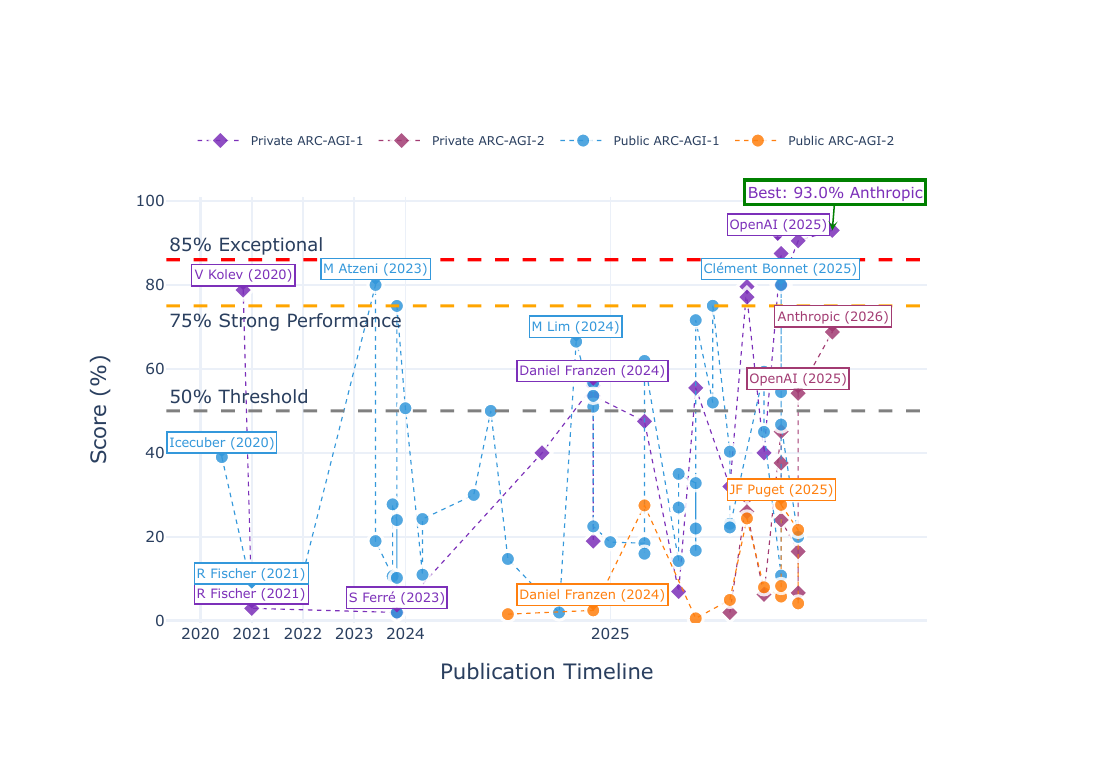}
\caption{
        ARC breakthroughs filtered for evaluation rigor
        ($\geq$100 tasks for ARC-AGI-1, $\geq$50 for ARC-AGI-2;
        see exclusion criteria in the Supplementary Material, Section~2).
        Circle markers denote public evaluations; diamond markers
        denote semi-private evaluations.
        The 2020--2023 timeline is compressed to emphasize the
        rapid progress in 2024--2026.
        On ARC-AGI-1, the rigorous maximum rose from 39.0\%
        (Icecuber, 2020, public) to 93.0\% (Opus~4.6, 2026,
        semi-private). ARC-AGI-2, introduced in late 2024, has
        seen its best rigorous score reach 68.8\% (Opus~4.6, 2026).
}
\label{fig:real_breakthroughs}
\end{figure}

Figure~\ref{fig:real_breakthroughs} shows that many highly-cited >80\% results
derive from <10 task evaluations. When restricted to $\geq$100 tasks, only six systems achieve $\geq$75\%: Huang (75\%/400), Pang (77.1\%/100), Kolev (78.8\%/100), Berman (79.6\%/100) Bonnet (80\%/100), and Google Gemini (87.5\%/100).

\begin{figure}[!t]
\centering
\includegraphics[width=0.90\textwidth, trim= 10 20 10 90, clip, scale=1.01]{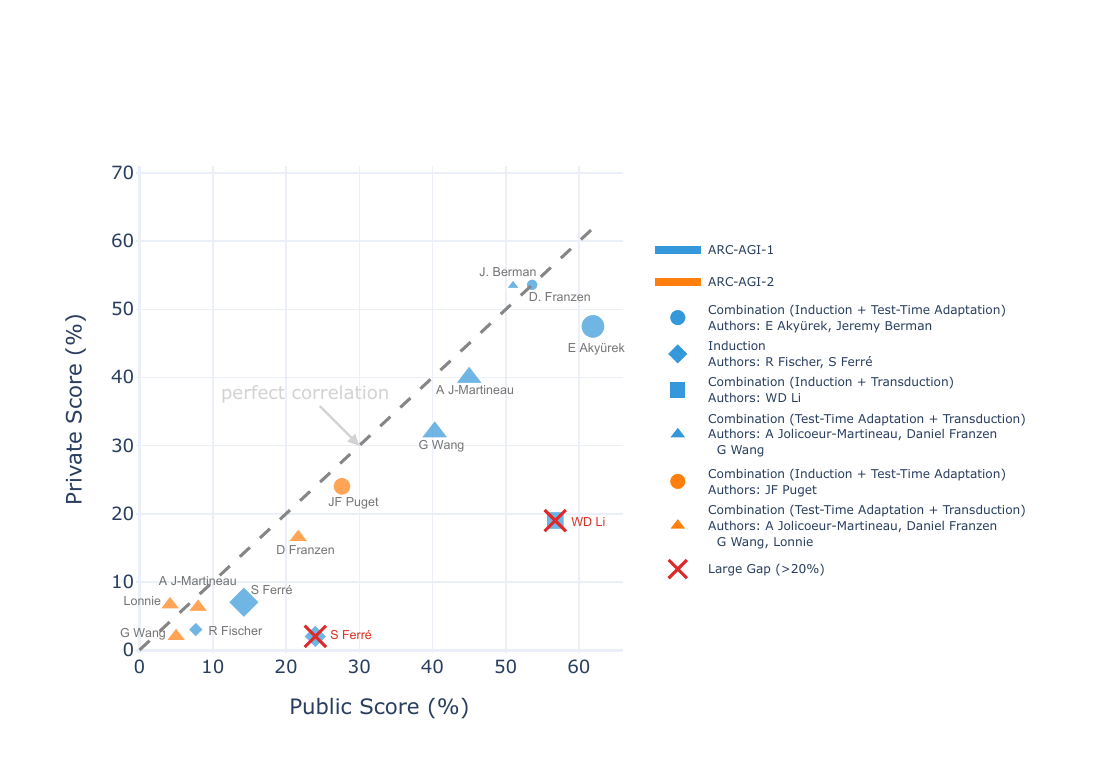}
\caption{
        Public versus semi-private ARC scores. Only 11 of 82
        papers report both (87\% transparency gap). On ARC-AGI-1, the
        mean public-to-semi-private drop is 10.8\% (median 6.1\%),
        with gaps ranging from $-$5.0\% to 37.8\%. On ARC-AGI-2, gaps
        are smaller (mean 2.0\%), likely because the semi-private set
        was introduced recently and fewer systems have been evaluated
        on both splits. Points above the diagonal indicate higher
        semi-private than public scores.
}
\label{fig:public_vs_private}
\end{figure}

The public-private gap (Figure~\ref{fig:public_vs_private}) shows 10--20\% drops,
but only 13 of 66 papers report both, a 80\% transparency gap that likely
underestimates the true generalization problem.

Methodological variance further complicates comparisons: papers report "accuracy"
vs "percentage solved" vs "best of N" with inconsistent N, use varying evaluation
sets (standard 400-task vs custom subsets), and apply different test-time
adaptation practices without consistent reporting.

Competition-verified results provide the most reliable estimates through independent
evaluation, hidden test sets, and standardized budgets. The ARC Prize competitions
exemplify this approach, though even competition results may not generalize to
harder variants. Improving evaluation transparency is as important to AGI progress
as algorithmic innovations.

\subsection{Evolutionary Case Studies}
\label{subsec:evolutionary-case-studies}

Examining how top-performing systems evolved reveals three paradigm-level
trajectories (detailed case studies are provided in the Supplementary Material).

\textbf{Program synthesis: from code to language to libraries.}
The representation strategy shifted from Python code generation
(Ouellette, 79.3\%) to natural language instruction evolution (Berman,
79.6\% at \$8.42/task) to library-based compositional learning (Pang, 77.1\%
at \$3.97/task). Berman's multi-agent architecture evolves 40 candidate
instructions per task using specialized generation, testing, and revision
agents. Pang's wake-sleep architecture accumulates a persistent program
library (538 programs from 1,000 training tasks), reducing LLM calls from
36 to 10 per task through cross-task knowledge transfer. Despite these
representational advances, all three show similar 2.5--3$\times$ degradation on
ARC-AGI-2 (Berman: 29.4\%, Pang: 26.0\%), indicating that representation
choice alone cannot overcome compositional reasoning limitations.

\textbf{Neural approaches: from pattern matching to test-time adaptation.}
Early neural systems achieved $<$5\%, but test-time fine-tuning transformed
the paradigm. MindsAI progressed from 5\% zero-shot to 58\% through
test-time fine-tuning combined with Augment Inference Reverse-Augmentation
and Vote (AIRV), a 680\% improvement validating that neural networks possess
latent reasoning capabilities unlocked by task-specific adaptation.
The LLM ARChitect system achieved 72.5\% under strict Kaggle constraints
(2$\times$T4 GPUs, 12 hours) through systematic optimization: tokenization
reduction (120K$\rightarrow$64 tokens), D8 augmentation, and batch DFS decoding.
However, neural approaches show the sharpest ARC-AGI-2 degradation (Wang:
40.3\%$\rightarrow$5\%; ARChitect: 72.5\%$\rightarrow$2.5\%), confirming that pattern
matching cannot handle increased compositional depth.

\textbf{Compute-intensive frontier: from o3 to GPT-5.2.}
The 390$\times$ efficiency gain from o3 (\$4,500/task, 87.5\%) to GPT-5.2 Pro
(\$11.64/task, 90.5\%) reflects reduced parallelism rather than algorithmic
breakthroughs. The performance cliff persists: GPT-5.2 Pro drops to 54.2\%
on ARC-AGI-2, confirming that neither scaling nor year-over-year improvements
resolve compositional generalization.

Across all trajectories, every system exceeding 70\% on ARC-AGI-1 employs
test-time adaptation, generates multiple hypotheses, and uses systematic
augmentation. Yet the universal ARC-AGI-2 cliff (Berman 79.6\%$\rightarrow$29.4\%,
Pang 77.1\%$\rightarrow$26.0\%, Wang 40.3\%$\rightarrow$5\%) and ARC-AGI-3 failure
(best: 12.58\%) indicate shared fundamental limitations.

\subsection{ARC Prize 2025: Refinement Loops as the Central Innovation}
\label{subsec:arc-prize-2025-winners}

The ARC Prize 2025 competition~\cite{arcprize2025_results} revealed a unifying
theme: \emph{refinement loops}---iterative cycles of generation, verification,
and correction. Table~\ref{tab:arc-prize-2025-winners} summarizes the winning
approaches; detailed architectural descriptions are provided in the
Supplementary Material.

\begin{table}[t]
\centering
\small
\setlength{\tabcolsep}{3pt}
\renewcommand{\arraystretch}{1.15}
\begin{tabular}{p{0.13\linewidth} r p{0.30\linewidth} p{0.38\linewidth}}
\hline
\textbf{System} & \textbf{Score} & \textbf{Key innovation} & \textbf{Refinement mechanism} \\
\hline
NVARC~\cite{nvarc2025_kaggle} (1st) & 24.0\% &
  Synthetic data mixing + 3-component ensemble (Qwen3 + TRM) &
  Data-generation loop: 266K puzzles from concept mixing; TRM hypothesis cycles; multi-component voting \\
ARChitects~\cite{the_architects_2025_techical_report} (2nd) & 16.5\% &
  LLaDA-8B masked diffusion LM with soft masking &
  102 recursive refinement steps via selective re-masking of uncertain regions \\
MindsAI~\cite{mindsai2025_kaggle} (3rd) & 12.6\% &
  TTFT + AIRV on Llama 2 70B &
  Task-specific fine-tuning + augmentation-based ensemble voting \\
\hline
Land~\cite{land2025_arc_solver} (public SOTA) & 72.9\% &
  Cross-model ensemble (GPT-5.2, Gemini~3, Opus~4.5) &
  Multi-model program synthesis + evaluator-based adjudication \\
\hline
\end{tabular}
\caption{ARC Prize 2025 winners (private ARC-AGI-2 test) and public
    leaderboard SOTA. All top systems implement refinement at different
    abstraction levels.}
\label{tab:arc-prize-2025-winners}
\end{table}

Refinement operates at multiple abstraction levels across these systems:
\emph{data-generation refinement} (NVARC's synthetic pipeline produces 103K
valid puzzles from concept mixing, validated against formal specifications),
\emph{inference-time refinement} (LLaDA-8B's 102 recursive masking steps;
TRM's nested hypothesis cycles), and \emph{ensemble refinement} (MindsAI's
AIRV; NVARC's multi-component voting). The ARChitects' masked diffusion
paradigm is treating output as a field to be
iteratively denoised rather than a sequence generated left-to-right, it
enforces global coherence through repeated consistency checks---a
qualitatively different inductive bias from autoregressive generation.

Land's cross-model ensemble~\cite{land2025_arc_solver} distributes tasks across three frontier models under varied
prompting configurations, eliciting executable programs validated in a sandbox,
then ranking candidates via evaluator models. This achieves 94.5\% on
ARC-AGI-1 (\$11.4/task) and 72.9\% on ARC-AGI-2 (\$38.9/task), suggesting
that inter-model diversity captures complementary reasoning strategies that
intra-model sampling alone cannot.

As the ARC Prize organizers noted, ``refinement is intelligence'': the
capacity for self-correction through feedback appears more central to fluid
intelligence than raw pattern recognition. However, refinement imposes
computational costs, and the challenge remains developing mechanisms that
approach human cognitive flexibility: fluid, low-overhead, and efficient.

\begin{rqbox}{Key Empirical Findings.}
    Unconstrained systems now reach 96.0\% on ARC-AGI-1 and 84.6\% on
    ARC-AGI-2, but at \$7--14 per task and with opaque training
    pipelines. Under Kaggle resource constraints, the best systems
    achieve only 24\% on ARC-AGI-2~\cite{arcprize2025_results, arcprize2025_leaderboard}.
    Refinement loops emerged as the central innovation, yet the
    30--60 percentage point gap between constrained and unconstrained
    regimes suggests that current progress reflects computational
    investment as much as architectural insight
    (Section~\ref{subsec:remaining-challenges}).
\end{rqbox}

\section{Discussion and Implications}\label{sec:discussion-and-implications}

The findings and insights presented in the previous sections
reveal both remarkable progress and fundamental limitations in current approaches to
abstract reasoning.
The findings suggest three questions that structure the closing discussion:
\begin{enumerate}
  \item \textbf{Measuring vs.\ understanding intelligence:}
  What does ARC-AGI reveal about measuring intelligence versus explaining intelligence?
  \item \textbf{Scaling vs.\ conceptual compression:}
  Why do current systems achieve performance primarily through resource scaling rather than conceptual compression?
  \item \textbf{Path to AGI:}
  What challenges remain on the route toward artificial general intelligence?
\end{enumerate}

\subsection{From Measurement to Mechanism (RQ1: Measuring vs.\ Understanding Intelligence)}
\label{subsec:rq1}

\paragraph{From comparative ranking to mechanistic explanation.}
The history of intelligence assessment reflects a gradual shift from \emph{ranking} to \emph{explaining}.
Early psychometric traditions, exemplified by ~\cite{binet_methodes_1905, spearman_general_1904},
were designed to quantify individual differences and enable comparative evaluation.
Over the past century, however, cognitive science has
increasingly emphasized identifying the architectural principles that make intelligent behavior
possible in the first place. A canonical example is Spelke’s \emph{core knowledge} proposal: humans
appear to share domain-general developmental priors for object representation, number, geometry,
and agency that support rapid learning across cultures ~\cite{spelke_core_2007}. This trajectory
highlights a crucial point: measurements become scientifically meaningful when they constrain
\emph{mechanisms}.

\paragraph{ARC-AGI as a computational probe of generalization.}
ARC-AGI represents a further step in this evolution. Rather than primarily asking how well a system
performs relative to others, ARC-AGI is best read as a probe of \emph{how} a system generalizes from
minimal evidence. In this sense, the benchmark asks not ``how smart is this system?'' but ``what kind
of intelligence does this system possess?'' This framing matters because the same score can be
achieved by qualitatively different routes. A system that reaches high accuracy by exploiting
benchmark-specific regularities or memorizing training patterns exhibits a different competence
profile than a system that extracts reusable primitives and composes them to solve novel tasks.
To answer RQ1, we argue that ARC-AGI scores should be interpreted as evidence of \emph{understanding}
only insofar as they are accompanied by mechanism-relevant context. Concretely, performance metrics
become meaningful when reported together with the following axes presented in Table \ref{tab:rq1_axes}.

\begin{table}[h!]
\centering
\small
\setlength{\tabcolsep}{4pt}
\renewcommand{\arraystretch}{1.15}
\begin{tabularx}{\linewidth}{>{\bfseries}p{0.17\linewidth} X}
\hline
\textbf{Axis} & \textbf{Interpretation question} \\
\hline
Efficiency &
What computational resources are required per solved task (e.g., sampling budget, number of refinement steps,
wall-clock time, monetary cost)? A method that succeeds only through massive test-time computation may
approximate intelligent behavior without revealing a compact mechanism. \\
\hline
Adaptability &
What changes at test time, and how? Does the system perform test-time adaptation (e.g., puzzle-specific training,
search, hypothesis revision, tool use), and is that adaptation principled or primarily brute-force? \\
\hline
Compositional scope &
Does the strategy generalize beyond the easiest regime? In particular, does it preserve competence as tasks demand
deeper multi-step composition (e.g., ARC-AGI-2) and, eventually, interactive learning (ARC-AGI-3)? \\
\hline
Error signature &
When the system fails, does it fail in interpretable, systematic ways that illuminate a missing primitive or decomposition
step, or does it collapse unpredictably across task families? \\
\hline
\end{tabularx}
\caption{Mechanism-relevant axes for interpreting ARC-AGI scores (RQ1).}
\label{tab:rq1_axes}
\end{table}

This checklist clarifies why ``higher score'' is not synonymous with ``more understanding.'' For example,
a system achieving 80\% accuracy through memorization demonstrates fundamentally different capabilities
than one reaching 60\% by discovering compositional primitives that transfer to unseen tasks. The former
is primarily a measurement of coverage; the latter is evidence of a mechanism that may scale in scope.
This mechanistic lens resonates with Leibniz’s philosophical criterion for distinguishing genuine
understanding from sophisticated pattern matching: one must observe whether a system can extract
abstract principles from minimal data, rather than merely exhibiting complex internal motion
~\cite{leibniz_monadology_1989}. Modern ARC-AGI systems operationalize this tension: they often appear intelligent
in output space while leaving open whether the internal process is compressive abstraction or
compute-amplified search.
The empirical record in Section~5 illustrates why this interpretation matters.
Current approaches span a wide cost--performance spectrum, from compute-intensive test-time adaptation to comparatively
efficient program synthesis.
Across the strongest systems, a unifying architectural motif has emerged:
\emph{refinement loops}.
This includes iterative cycles of generation, verification, and correction which consistently
outperform single-pass prediction. The ARC Prize 2025 results reinforce this conclusion, suggesting that
self-correction through feedback is not a peripheral trick but a central capability in high-performing
ARC-AGI systems.
At the same time, the persistent performance cliff from ARC-AGI-1 to ARC-AGI-2 indicates that current
refinement-based mechanisms do not yet constitute robust compositional generalization.
The cliff appears
across paradigms and implementations, implying a limitation that is architectural rather than merely
a question of ``more compute.''

\begin{rqbox}{Bridge to RQ2.}
Taken together, these observations suggest a concrete interpretation of RQ1: ARC-AGI is valuable not
because it produces a single score, but because it forces us to ask \emph{what was paid} (efficiency),
\emph{what was adapted} (adaptability), and \emph{what actually generalized} (compositional scope).
This sets up RQ2: if the best results often come from spending more resources at test time, why does
scaling so reliably improve performance, and what does that imply about the absence of conceptual
compression in current systems?
\end{rqbox}

\subsection{The Compression Paradox: (RQ2: Scaling vs.\ Conceptual Compression)}
\label{subsec:compression-paradox}

A striking pattern emerges across all high-performing systems documented in
Section~\ref{sec:empirical-analysis}.
Current AI achieves performance through scaling computational resources rather than discovering compressed representations.
This ``more with more'' paradigm, increasing capabilities by increasing resources,
contrasts sharply with human intelligence, which operates on a ``more with less''
principle of solving increasingly diverse problems through parsimonious conceptual
frameworks, see Table \ref{tab:rq2_morewithmore}.
This distinction, recently formalized by Krakauer, and Mitchell
through complexity science~\cite{krakauer_emergence_2025}, illuminates fundamental
limitations in current approaches to artificial general intelligence.

\begin{table}[h!]
\centering
\small
\setlength{\tabcolsep}{4pt}
\renewcommand{\arraystretch}{1.15}
\begin{tabularx}{\linewidth}{>{\bfseries}p{0.17\linewidth} X X}
\hline
 \textbf{Dimension} & \textbf{``More with more'' (scaling)} & \textbf{``More with less'' (compression)} \\
\hline
Primary driver &
More samples, more compute, more data, larger ensembles &
Effective theories: coarse-grained abstractions that screen off irrelevant detail~\cite{krakauer_emergence_2025} \\
\hline
What transfers &
Surface regularities and task-family coverage &
Structural principles that generalize across substrates (analogy)~\cite{hofstadter_analogy_2001} \\
\hline
Typical failure &
Cliff-like breakdown under deeper composition or distribution shift &
Graceful degradation: errors reveal missing primitives/decompositions \\
\hline
Interpretation &
Capability accumulation (coverage/search) &
Emergent intelligence (compression, explanation, reuse)~\cite{krakauer_emergence_2025} \\
\hline
\end{tabularx}
\caption{Two routes to performance. ARC-AGI progress is currently dominated by scaling, while human-like fluid intelligence is associated with conceptual compression.}
\label{tab:rq2_morewithmore}
\end{table}

\subsubsection{Intelligence as Emergent Compression}

Krakauer et al.\ distinguish \emph{emergent capabilities} (specific functions excelling
at particular tasks) from \emph{emergent intelligence} (the capacity to discover
coarse-grained representations enabling broad problem-solving through analogical
reasoning)~\cite{krakauer_emergence_2025}. True intelligence, they argue, manifests
through discovering ``effective theories'': compressed representations that screen off
irrelevant details while preserving predictive power across contexts---much as the ideal
gas law predicts macroscopic behavior without tracking individual molecules, or as humans
recognize that rotation principles apply identically to mental imagery and physical
manipulation. Intelligence thus lies not in accumulating specialized functions but in
discovering minimal principles with maximal explanatory scope, precisely what current
AI systems struggle to achieve on ARC-AGI benchmarks.

\subsubsection{The Energy Efficiency Criterion as the diagnostic not just an engineering metric}

The human brain operates on approximately 20 watts, less than a modern LED light
bulb, yet consistently outperforms megawatt-consuming AI systems on ARC-AGI tasks.
This disparity is not merely impressive engineering but reflects a fundamental
principle: genuine intelligence involves finding minimal-energy paths through
problem spaces~\cite{krakauer_emergence_2025}. Efficiency serves as a criterion
for intelligence because discovering compressed representations inherently requires
less computational work than exhaustive search or memorization.

This efficiency criterion illuminates the cost-performance tradeoffs documented
in Section~\ref{subsec:cost-performance-frontiers} and Table~\ref{tab:cost_perf_narrow}.
Despite a 390$\times$ cost reduction from o3 to GPT-5.2 Pro, this improvement largely
reflects reduced parallelism rather than more efficient reasoning algorithms. The
fundamental pattern persists: performance on ARC-AGI-2 drops to 54.2\% even for
frontier models, confirming that efficiency gains do not resolve compositional
generalization limitations.

Humans solve ARC tasks in seconds with approximately 2,400 joules per task, while
AI systems remain orders of magnitude less efficient. As Chollet observed: ``While
ARC 1 is now saturating, SotA models are not yet human-level on an efficiency
basis''~\cite{arcprize2025_leaderboard}. This gap indicates current approaches achieve
performance through parallelism and search rather than discovering compressed
representations.

The ARC Prize 2025 analysis~\cite{arcprize2025_results} crystallizes this distinction:
the \emph{accuracy gap} to human performance is now ``primarily bottlenecked by
engineering'' (solvable through sufficient compute, data coverage, and verifiable
feedback) while the \emph{efficiency gap} remains ``bottlenecked by science and
ideas.'' Current AI reasoning capability is fundamentally \emph{knowledge-bound}:
performance depends on domain coverage in training data plus verifiable task
feedback. Human reasoning, by contrast, is \emph{knowledge-free} in Chollet's
sense, capable of generalizing abstract principles to domains never encountered.
This asymmetry explains why scaling improves accuracy on well-covered domains
(mathematics, coding, even ARC-AGI-1) while efficiency remains static: more
parameters compress more knowledge but do not discover the compression mechanisms
themselves. Closing the efficiency gap requires separating knowledge from
reasoning, learning \emph{how to learn} rather than what to know.

The ARC Prize 2025 winners exemplify this knowledge-bound limitation: NVARC's
winning 24.03\% required 266K synthetic puzzles and 3.2M augmented samples,
while the ARChitects' 16.53\% demanded 39 hours on 8$\times$H100 GPUs, yet
these substantial resource requirements achieved only 16--24\% on puzzles humans solve effortlessly
(Section~\ref{subsec:arc-prize-2025-winners}). Notably, open-source solutions
enable full scrutiny of these pipelines, whereas proprietary models remain
opaque regarding training corpora and potential benchmark exposure, underscoring
the scientific value of open research for establishing genuine capability
baselines.
Taken together, scaling-driven performance, knowledge-bound reasoning, and
persistent efficiency gaps collectively define the compression paradox at the
heart of current ARC-AGI progress.



\begin{rqbox}{Bridge to RQ3.}
    Current approaches that achieve the strongest reported ARC-AGI results
    approximate aspects of abstraction discovery without fully achieving it.
    Program synthesis methods seek reusable primitives but struggle with
    semantic grounding. Neuro-symbolic systems attempt to bridge perception and
    symbols but rely on hand-engineered ontologies. Test-time adaptation explores
    hypothesis spaces through search rather than principled abstraction.
    RQ2 suggests that the current progress is largely driven by scaling
    and refinement rather than by discovering compact, transferable abstractions.
    The remaining question is therefore mechanistic: \emph{which missing components}
    (composition, grounding, interaction) prevent ``more with less'' generalization,
    and what architectural directions could close the efficiency gap rather
    than only the accuracy gap?
\end{rqbox}

\subsection{Remaining Challenges on the Path to AGI}
\label{subsec:remaining-challenges}

Despite substantial progress on ARC-AGI-1, performance on compositionally complex
variants remains far below human baselines, exposing three critical challenges that
current approaches have not resolved.

\subsubsection{The Compositional Generalization Bottleneck}
Compositional generalization, the ability to systematically recombine known primitives
in novel ways, represents the binding constraint on current approaches. Systems achieving
70-80\% on ARC-AGI-1 consistently drop to 20-30\% on ARC-AGI-2 and below 15\% on
ARC-AGI-3~\cite{chollet_arc_2025, chollet_arc-agi-2_2025}. This is not gradual
degradation but catastrophic failure once compositional depth exceeds system capacity. The challenge manifests through multiple failure modes. First, combinatorial explosion
in search spaces makes deeper compositions intractable. ConceptSearch generates 400
candidates for two-step transformations but 8,000 for three-step transformations~\cite{singhal_conceptsearch_2025}.
Without mechanisms to prune search spaces through principled abstraction, the exponential
growth of hypothesis spaces overwhelms computational budgets. Second, insufficient
learning signal plagues gradient-based approaches. MindsAI achieves strong ARC-AGI-1
results but admits that three-step reasoning provides inadequate training signal for
their learning mechanisms. The distribution shift from ARC-AGI-1 to ARC-AGI-2 is not
merely quantitative but qualitative, requiring generalization along compositional
dimensions absent from training data.

Third, systems exhibit brittle failure modes rather than graceful degradation. Humans
confronting difficult compositional tasks attempt solutions even when uncertain,
exhibiting partial success and systematic errors that reveal underlying reasoning
strategies. Current AI systems show catastrophic failure: ANPL drops from 75\% on
ARC-AGI-1 to near-zero on harder variants with no intermediate attempts or partial
solutions. This brittleness suggests fundamental differences in how humans and current
AI represent compositional structure.
Addressing compositional generalization likely requires mechanisms for hierarchical
decomposition rather than sequential chaining. Current systems treat composition as
concatenation, searching over all possible sequences of operations. Human reasoning
employs hierarchical decomposition: breaking complex transformations into manageable
sub-problems, solving these with appropriate abstractions, then composing solutions
through structured interfaces. This hierarchical approach scales sub-exponentially
because it reduces effective search space at each level. No current architecture
implements explicit mechanisms for such hierarchical reasoning, explaining the
consistent performance cliff across all paradigms.

\subsubsection{Symbol Grounding for Learned Primitives}
Program synthesis approaches discover transformation primitives, but these primitives
often lack the semantic grounding that makes human concepts flexible and transferable.
A human understanding of ``rotation'' applies to mental imagery, physical objects,
abstract diagrams, and temporal sequences. Current AI primitives, by contrast, remain
tied to specific instantiations, limiting their compositional utility.

This manifests as the classical symbol grounding problem~\cite{harnad_symbol_1990}:
how do abstract symbols acquire meaning beyond formal manipulation? Symbolic AI
historically suffered from symbols manipulated according to syntactic rules without
semantic content, as Searle's ``Chinese Room'' critique noted~\cite{searle_minds_1980}. Neural
networks achieve perceptual grounding but lose compositional structure. Neuro-symbolic
approaches attempt to bridge this gap, but
current implementations rely on hand-engineered primitive vocabularies rather than discovered
abstractions.

A genuine solution would enable systems to recognize when different surface forms
instantiate the same abstract operation and compose primitives in novel ways not
seen during training. Ouellette et al.'s neural-guided synthesis achieves 79.3\%
on ARC-AGI-1~\cite{ouellette_towards_2024} by learning program representations,
but discovered primitives function as optimized parameters rather than understood
concepts: they cannot be explained or transferred to novel domains. Promising
directions include compositional abstraction discovery~\cite{ellis_dreamcoder_2021}
and concept learning frameworks~\cite{lake_building_2017}, but no current system
demonstrates robust symbol grounding at the flexibility required for human-level
abstract reasoning.

\subsubsection{Architectural Paths Forward}
The ARC Prize 2025 Paper Awards~\cite{arcprize2025_results} illuminate unexplored
directions that diverge from mainstream scaling approaches. Three innovations merit
particular attention.
Table \ref{tab:rq3_paths_simple} summarizes these
representative directions.

\begin{table}[h!]
\centering
\small
\setlength{\tabcolsep}{4pt}
\renewcommand{\arraystretch}{1.15}
\begin{tabularx}{\linewidth}{>{\bfseries}p{0.18\linewidth} Y}
\hline
\textbf{Direction} & \textbf{Summary} \\
\hline

Recursive latent refinement &
Jolicoeur-Martineau's Tiny Recursive Model (TRM) achieves 45\% on ARC-AGI-1 and 8\% on ARC-AGI-2 with only
7M parameters, less than 0.01\% of leading LLMs~\cite{jolicoeur2025_trm}. The architecture alternates between
``think'' steps ($z \leftarrow f(x,y,z)$) that accumulate latent reasoning and ``act'' steps ($y \leftarrow g(y,z)$)
that refine answers, unrolled across 16 iterations. This demonstrates that \emph{recursive depth compensates for
parameter count}: computational depth of 42 effective layers through iteration enables generalization that massive
parameter counts alone cannot achieve. \\
\hline

Self-improving program synthesis &
Pourcel et al.'s SOAR framework fine-tunes LLMs on their own evolutionary search traces, achieving 52\% on
ARC-AGI-1 without hand-engineered DSLs~\cite{pourcel-2025-self-improving}. The key insight is that failed solution
attempts contain valuable learning signal. By converting search failures into training data, the system discovers that
``positive transfer between sampling and refinement tasks'' enables progressive improvement across iterations. This
suggests that learning \emph{how to search} may matter more than searching with fixed capabilities. \\
\hline

Compression as intelligence &
Liao and Gu's CompressARC solves 20\% of ARC-AGI-1 with 76K parameters and \emph{zero pretraining}, optimizing
description length purely at inference time~\cite{liao2025_compressarc}. The approach formalizes puzzle-solving as
code-golfing: finding the shortest program encoding correct solutions. This operationalizes the Solomonoff--Kolmogorov
thesis that compression \emph{is} intelligence~\cite{solomonoff_formal_1964}, demonstrating that MDL-based inference
can substitute for massive pretraining. \\
\hline

\end{tabularx}
\caption{Mechanism-oriented directions highlighted by ARC Prize 2025 paper awards.}
\label{tab:rq3_paths_simple}
\end{table}

\paragraph{Toward Self-Grounding Primitives.}
These innovations reveal a progression: 2024's breakthrough was \emph{test-time
adaptation}, adapting at inference rather than training. 2025's theme is
\emph{refinement loops}, iterative correction cycles. The logical 2026 extension
may be \emph{recursive abstraction discovery}, systems that discover their own
primitive vocabulary during inference, grounding symbols through compression rather
than pretraining. The key challenge remains bridging perceptual grounding with
compositional structure: TRM achieves efficient recursion but lacks symbolic
interpretability; SOAR learns search strategies but requires LLM foundations;
CompressARC grounds through compression but within fixed architectural primitives.
A synthesis enabling \emph{self-grounding primitives}, discovered, compressed,
and compositionally recombined within a single inference episode, would represent
the next frontier. ARC-AGI-3's interactive paradigm compounds this challenge by requiring that such primitives also support world model induction through
exploration~\cite{Ying2025-WorldModels}.

\begin{rqbox}{What ARC-AGI Reveals About AGI.}
The benchmark's evolution shifts the question: frontier models nearly reach human baselines on ARC-AGI-1, but at $\sim$20--40$\times$ human cognitive effort, with opaque training, and they degrade sharply on ARC-AGI-2/3. Two paths remain: \textbf{Scaling} (more parameters/test-time compute for brute-force coverage---feasible but costly and opaque) or \textbf{Innovation} (new mechanisms for compositional abstraction and efficient search). Evidence suggests neither alone is enough; the best approaches blend learned representations, explicit composition, and efficient search. ARC-AGI is thus more diagnostic than pass/fail---the key is not \emph{whether} AI matches humans, but \emph{how} it does so and what that implies for AGI.

\end{rqbox}

\section{Related Work}\label{sec:related-work}

This section situates our survey within the broader landscape of AGI research,
examining theoretical foundations, alternative benchmarks, and prior documentation
of ARC-AGI approaches. Zinkevich~\cite{zinkevich_arc_slr_2025} presents a PRISMA-guided
systematic literature screening 538 manuscripts and classifying 62 ARC Prize 2024
approaches via technique tagging and ensemble synergy analysis using task-level
performance data. Our survey extends this foundation with full temporal coverage
(2019-2025), cross-generation analysis spanning ARC-AGI-1, ARC-AGI-2, and ARC-AGI-3,
detailed taxonomic descriptions of four paradigms, cost-performance frontier analysis,
and results from the ARC Prize 2025 competition. Prior foundational work includes Chollet's
original benchmark paper~\cite{chollet_measure_2019}, the ARC Prize technical
reports~\cite{arcprize2025_results}, and position papers on AGI
measurement~\cite{steinbauer_position_2025}.

\subsection{Theoretical Frameworks for AGI and Intelligence Measurement}
\label{subsec:theoretical-frameworks}

Chollet's 2019 paper ``On the Measure of Intelligence''~\cite{chollet_measure_2019}
provided the theoretical foundation for ARC-AGI by defining intelligence as
\textit{skill-acquisition efficiency}, emphasizing scope, generalization difficulty,
and priors rather than raw performance. This framework directly challenges
benchmarks that reward capability accumulation without considering data efficiency.
The cost-performance analysis in Section~\ref{sec:empirical-analysis} validates
this insight: systems with similar accuracy exhibit efficiency gaps exceeding
three orders of magnitude.

Steinbauer et al.~\cite{steinbauer_position_2025} propose six pillars for
efficient general intelligence: compositional representations, systematic
generalization, causal reasoning, abstract concepts, meta-learning, and continual
learning. Our empirical analysis validates several predictions: hybrid approaches
combining program synthesis with test-time adaptation demonstrate that compositional
representations and meta-learning are necessary for high performance, while
revealing that current implementations remain insufficient for robust generalization
across ARC-AGI versions.

\subsection{Alternative AGI and Reasoning Benchmarks}
\label{subsec:alternative-benchmarks}

ARC-AGI exists within a broader ecosystem of benchmarks attempting to measure
aspects of AGI and reasoning capability. Understanding ARC-AGI's
relationship to these alternatives clarifies its unique contributions and limitations.

\textbf{Broad AGI Evaluation Suites:} Big-Bench~\cite{srivastava_beyond_2023}
and MMLU~\cite{hendrycks_measuring_2021} assess language models across hundreds
of diverse tasks spanning mathematics, science, history, and common sense reasoning.
These benchmarks measure breadth of acquired knowledge and capabilities, operating
in the complementary regime to ARC-AGI's focus on few-shot generalization from
minimal data. Large language models achieve >90\% on many MMLU categories yet
remain below 20\% on ARC-AGI-2, demonstrating that knowledge breadth and fluid
reasoning represent orthogonal challenges. Our survey focuses exclusively on
the latter.

\textbf{Visual Reasoning Benchmarks:} RAVEN's Progressive Matrices~\cite{zhang_raven_2019}
and similar abstract visual reasoning tasks share surface similarity with ARC-AGI's
grid-based format. However, RAVEN provides multiple-choice answers and emphasizes
pattern recognition over transformation synthesis. While some ARC-AGI techniques
transfer to these domains, the open-ended generation requirement in ARC-AGI
creates fundamentally different solution constraints.

\textbf{Human Baseline Studies:} H-ARC~\cite{legris_h-arc_2024} provides robust
human performance estimates through a large-scale study collecting over 1,700
solution attempts from 316 participants on all 400 ARC-AGI-1 public evaluation
tasks. Their findings reveal that task difficulty for humans does not correlate
with difficulty for AI systems, suggesting fundamentally different reasoning
strategies. This disconnect reinforces ARC-AGI's value as a diagnostic tool:
high AI performance on tasks humans find difficult, combined with AI failure
on tasks humans find easy, indicates reliance on surface features rather than
the abstract reasoning humans employ.

\textbf{Interactive and Embodied Environments:} BabyAI~\cite{chevalier-boisvert_babyai_2019},
NetHack~\cite{kuttler_nethack_2020}, and Crafter~\cite{hafner_benchmarking_2021}
evaluate sequential decision-making and long-horizon planning in interactive
environments. ARC-AGI-3's shift toward interactive evaluation represents convergence
with this paradigm, emphasizing \emph{world model induction}, the capacity to
rapidly construct and refine internal environment representations through
exploration~\cite{Ying2025-WorldModels}. AutumnBench~\cite{autumnbench2025}
provides a complementary evaluation framework, testing reward-free discovery
of environment dynamics through masked prediction, planning, and change detection.
Both benchmarks reveal that frontier LLMs consistently underperform humans on
interactive reasoning, suggesting fundamental gaps in adaptive exploration
capabilities.

\textbf{Mathematical and Coding Benchmarks:} MATH~\cite{hendrycks_measuring_2021_math},
GSM8K~\cite{cobbe_training_2021}, HumanEval~\cite{chen_evaluating_2021}, and
MBPP~\cite{austin_program_2021} assess formal reasoning in symbolic domains.
Test-time adaptation techniques developed for ARC-AGI have shown transfer to these domains, with
search-based methods achieving state-of-the-art results on mathematics competitions.
This suggests ARC-AGI research advances fundamental reasoning capabilities beyond
visual pattern manipulation.

\textbf{Compositional Generalization:} SCAN~\cite{lake_generalization_2018} and
COGS~\cite{kim_cogs_2020} specifically test compositional generalization in
language understanding. While these benchmarks probe similar capabilities to
ARC-AGI, they operate in discrete symbolic domains with explicitly defined
grammars. ARC-AGI's challenge lies in discovering compositional structure from
perceptual inputs without explicit symbolic representations, a harder problem
requiring both perception and abstraction.

\section{Conclusion}\label{sec:conclusion}

This survey has traced the six-year journey of ARC-AGI research, analyzing over
66 approaches across three progressively challenging benchmark generations. The
findings reveal both remarkable progress (from near-zero to 90.5\% on ARC-AGI-1) and
persistent fundamental limitations that expose the gap between current AI and human
intelligence. We conclude by synthesizing key insights, examining what this gap
reveals about the nature of intelligence, and identifying the path forward toward
artificial general intelligence.

\subsection{The Human-AI Performance Gap}
\label{subsec:human-ai-gap}

The most striking finding is not how far AI has progressed but how far it remains
from human capabilities. Humans achieve near-100\% accuracy on all ARC-AGI versions
with minimal effort, while the best AI reaches 90.5\% on ARC-AGI-1, 54.2\% on
ARC-AGI-2, and 13\% on ARC-AGI-3~\cite{arcprize2025_results}. This trajectory
reveals three persistent dimensions of the human-AI gap (extended analysis in
Supplementary Material).

\textbf{The compositional generalization gap.} Performance degrades 2.5-3$\times$
per compositional step, consistently across all paradigms, indicating a shared
fundamental limitation rather than paradigm-specific weakness. Humans exhibit no
such degradation, suggesting they employ hierarchical decomposition rather than
the sequential chaining that causes AI systems to hit hard performance
cliffs (Section~\ref{subsec:remaining-challenges}).

\textbf{The efficiency gap.} Despite a 390$\times$ cost reduction from o3 to GPT-5.2
Pro, AI systems remain 20-40$\times$ less efficient than human cognition. This gap
is not a secondary engineering concern but a diagnostic criterion: genuine
intelligence discovers compressed representations rather than approximating results
through search (Section~\ref{subsec:compression-paradox}).

\textbf{The abstraction discovery gap.} Current systems succeed on ARC-AGI-1
through test-time search, accumulated libraries, or pattern matching within
learned distributions, but none demonstrate the spontaneous abstraction discovery
that allows humans to infer transferable principles from minimal
examples (Section~\ref{subsec:remaining-challenges}).

\noindent\textbf{Interpreting the gap.}
These three dimensions reflect a fundamental distinction between accumulated
capabilities and genuine intelligence~\cite{krakauer_emergence_2025,
chollet_measure_2019}: current systems achieve ``more with more'' (scaling
resources) rather than ``more with less'' (discovering compressed
representations). The cross-paradigm performance cliff from ARC-AGI-1 to
ARC-AGI-2 confirms that all current learning paradigms, whether gradient
descent, evolutionary search, or symbolic reasoning, optimize for capability
accumulation rather than intelligent compression.

\subsection{Emerging Paradigms: From Refinement Loops to World Models}
\label{subsec:emerging-paradigms}

The ARC Prize 2025 competition revealed \emph{refinement loops}, iterative cycles
of generation, verification, and correction, as the central innovation across
winning approaches~\cite{arcprize2025_results}. The Paper Awards highlighted three
promising directions: recursive latent refinement (TRM achieving 45\% with 7M
parameters), self-improving program synthesis (SOAR reaching 52\% by learning from
search traces), and compression-based inference (CompressARC solving 20\% with zero
pretraining)~\cite{jolicoeur2025_trm, pourcel-2025-self-improving, liao2025_compressarc}. These
approaches demonstrate that recursive depth, learning from failure, and compression
objectives can substitute for massive scale.

The transition to ARC-AGI-3 (interactive environments) signals a further shift:
best performance of 13\% versus 100\% human baseline exposes that capabilities
developed for passive pattern recognition do not transfer to active learning.
This aligns with emerging research on world models and interactive
intelligence~\cite{Ying2025-WorldModels}. Three directions emerge from this
convergence:

\noindent\textbf{From pattern matching to world modeling.} Genuine intelligence requires
building predictive models of environment dynamics rather than memorizing
input-output mappings. World models enable counterfactual reasoning, compositional
understanding, and efficient planning. ARC-AGI-3's~\cite{Ying2025-WorldModels} interactive format tests these
capabilities: agents must infer environmental mechanics through exploration and
discover goals from sparse feedback. The failure of ARC-AGI-1-optimized systems
to transfer to ARC-AGI-3 demonstrates that static reasoning misses essential
aspects of intelligence requiring active model building.

\noindent\textbf{From fixed architectures to meta-learning.} The success of hybrid approaches
suggests intelligence requires dynamic strategy
selection rather than fixed pipelines. Current hybrids use hand-engineered
orchestration; humans employ learned orchestration through meta-reasoning. The
challenge is developing hierarchical meta-learning architectures where higher-level
systems learn to orchestrate lower-level reasoning based on task
characteristics.

\noindent\textbf{From memorization to emergent compression.} General intelligence requires
mechanisms for emergent compression, spontaneous discovery of minimal principles
with maximal explanatory power. Current learning paradigms optimize for prediction
accuracy, incentivizing memorization. Achieving human-like intelligence likely
requires optimization objectives that explicitly reward compression. The 2025
Paper Awards validate this direction: CompressARC's MDL-based approach and TRM's
recursive refinement both achieve strong results through compression principles
rather than scale~\cite{liao2025_compressarc, jolicoeur2025_trm}.

\subsection{The Path Forward}
\label{subsec:path-forward}

The empirical evidence and theoretical analyses converge on a clear conclusion:
achieving artificial general intelligence requires architectural innovations beyond
scaling current paradigms. Two paths forward emerge, neither fully satisfactory
alone but potentially complementary.
\begin{itemize}
    \item \textbf{Path 1: Brute-force scaling.} High-compute test-time approaches demonstrate
that sufficient computation can solve ARC-AGI-1 tasks. However, three limitations
constrain this path: exponential scaling requirements make approaching human
performance economically prohibitive; compositional complexity creates hard barriers
(performance drops sharply on ARC-AGI-2 despite massive compute); and the efficiency
gap suggests this path produces capability without understanding.

    \item \textbf{Path 2: Architectural innovation.} Alternative approaches emphasize
discovering new mechanisms for compositional abstraction, symbol grounding, and
emergent compression. Efficient systems demonstrate that performance need not
require massive computation when architectures discover appropriate abstractions.
The 2025 Paper Awards (TRM, SOAR, CompressARC) validate this direction, achieving
competitive results through recursive refinement, self-improving search, and
compression objectives rather than scale.
\end{itemize}

The most promising path integrates
insights from both approaches, learned representations from pre-training, explicit
compositional mechanisms for systematic generalization, world models for interactive
learning, meta-learning for strategy discovery, and compression-based optimization.
The gap from current best performance to human baseline indicates that substantial
architectural innovation remains necessary.

\noindent\textbf{ARC-AGI as Diagnostic Instrument.}
ARC-AGI serves not as a goalpost to surpass but as a diagnostic instrument revealing
where current AI diverges from human intelligence. Each version exposes a different
capability dimension: basic abstraction extraction (ARC-AGI-1), deeper compositional
reasoning (ARC-AGI-2), and interactive world modeling (ARC-AGI-3).
The trajectory from 2\% (2019) to 90.5\% (2025) on ARC-AGI-1 demonstrates that
rapid progress is possible when challenges are well-designed. The 390$\times$ efficiency
improvement in a single year (o3 to GPT-5.2) shows that cost barriers can fall
rapidly. Yet the persistence of the compositional bottleneck across versions
suggests this progress reflects increasingly sophisticated pattern matching rather
than breakthroughs in compositional generalization.

Three methodological lessons emerge: (1) benchmark evolution proves essential: static
benchmarks lead to overfitting; (2) evaluation rigor determines measurement
accuracy, since comprehensive evaluations yield dramatically different results than
limited testing; (3) efficiency metrics provide diagnostic value beyond economics,
revealing algorithmic differences more clearly than performance metrics alone.

\subsection{Concluding Reflection}
\label{subsec:concluding-reflection}
The question at the heart of ARC-AGI research is not simply whether AI
can solve abstract reasoning puzzles, but what \emph{kind} of intelligence
contemporary systems instantiate. Across six years and 66 approaches,
the evidence suggests consistent patterns. Specifically, modern systems
can achieve impressive competence through refined pattern matching and
increasingly powerful search, yet they still struggle with the compositional
abstraction and emergent compression that make human reasoning fluid,
transferable, and efficient.

One way to see ARC-AGI's contribution is as a deliberately evolving
diagnostic instrument. ARC-AGI-1 asked whether systems can extract
a novel rule from a handful of examples and apply it to an unseen instance.
ARC-AGI-2 intensified the same question under greater compositional depth
and stricter resistance to overfitting. It revealed that gains on ARC-AGI-1
do not robustly translate to deeper compositions.
ARC-AGI-3 then shifts the axis entirely, and asks whether an agent
can acquire skills efficiently in a new environment through interaction.
This brings testing exploration, memory, planning, and the construction
of actionable internal models together, rather than static input-output induction.
Taken together, the versions progressively reduce the room for brute-force
substitution, from static generalization, to compositional generalization,
to interactive skill acquisition.

The 2025 innovations (including refinement loops, recursive latent reasoning,
self-improving search, and compression-based inference) show that progress
can come from methodological advances, not scale alone.
However, the gap to human baselines on ARC-AGI-2 persists, and early ARC-AGI-3
results point to the same limitation.
Systems are becoming stronger and more cost-efficient, yet transfer
remains fragile under deeper composition and interactive model building.

If ARC-AGI continues to evolve along its underlying thesis, that intelligence
is best revealed by \emph{skill-acquisition efficiency}, then ARC-AGI-4 will likely
push beyond acquiring skills in a single environment.
This possibility will be designed toward acquiring \emph{principles}
that transfer across environments. In this framing, ARC-AGI-4 is expected
to emphasize rapid causal abstraction from limited interaction,
and to evaluate whether such abstractions transfer reliably across
surface variation, tighter budgets, and novel contexts.
The natural next step, ARC-AGI-5 and its possible further versions,
would extend this into lifelong generalization.
On this trajectory, the central question is not whether AI will eventually solve ARC-AGI,
but \emph{how}. A solution achieved by ever-larger search and compute would signal
the continued scaling of capability.
A solution achieved by compact, compositional, transferable abstractions learned efficiently through interaction would signal progress toward intelligence in the stronger sense that ARC-AGI was built to diagnose.
The benchmark's future versions will therefore not only track performance, they are expected to increasingly clarify which computational principles, if any, can bridge the remaining distance between today's systems and human generality.

\vspace{0.5em}
\noindent In that sense, nature remains the most stringent reference point for general
intelligence, and the most informative source of constraints on what our models must
ultimately capture, and perhaps the clearest guide to mechanisms we have not yet
discovered:

\begin{quote}
    \emph{``Natural evolution suggests that AGI won't come from larger models
    that cram more and more specific knowledge, but from discovering the meta-rules
    that allow a system to grow and adapt its own architecture in response to the environment.''}
\begin{flushright}
---Fran\c{c}ois Chollet (@fchollet), 4 Feb, 2026 - X post~\cite{chollet2026}
\end{flushright}
\end{quote}

\bibliographystyle{ACM-Reference-Format}
\bibliography{
   references/arc-agi,
   references/agi,
   references/artificial-intelligence,
   references/psychology,
   references/philosophy
}
\end{document}